\documentclass[global,twocolumn]{svjour}    
\pdfoutput=1

\smartqed  
\usepackage{ifthen}
\usepackage{times}
\usepackage{mathptmx}      
\usepackage{url}
\usepackage{verbatim}
\usepackage{enumitem}
\usepackage{hhline}

\usepackage{mathtools}
\usepackage{color}
\usepackage{epstopdf}
\usepackage{tabularx}
\usepackage{multirow}
\usepackage[round]{natbib}

\usepackage[table]{xcolor}

\usepackage{url}
\usepackage[]{hyperref}

\usepackage{graphicx}
\usepackage{amsmath}
\usepackage{amssymb,amsmath}
\usepackage{comment}
\usepackage{xspace}
\usepackage{subfigure}
\usepackage{algorithm}

\definecolor{darkblue}{rgb}{0,0.1,0.5}

\usepackage{pifont}

\usepackage{booktabs}

\newcommand{\eg}{\emph{e.g.}\xspace}

\newcommand{\ie}{\emph{i.e.}\xspace}

\newcommand{\Tau}{\mathrm{T}}

\newcommand{\imemotion}{$iMotion$\xspace}

\newcommand{\tubeletvid}{$\Tau_{\text vid}$\xspace}

\newcommand{\tubeletime}{$\Tau_{\text iMotion}$\xspace}

\makeatletter     
\def\subsubsection{\@startsection{subsubsection}{3}{\z@}%
	{-21dd plus-4pt minus-4pt}{5.5dd plus 4pt minus4pt}{\normalsize\itshape}}    
\makeatother

\journalname{International Journal of Computer Vision} 

\graphicspath{{.}{./figs/}}

\renewcommand{\paragraph}[1]{\medskip {\noindent \bf #1}}

\DeclareMathOperator{\size}{\Gamma}

\graphicspath{{.}{./figs/}}

\def\ie{\emph{i.e.}\xspace}

\def\eg{\emph{e.g.}}

\begin{document}

\sloppy

\title{Tubelets: Unsupervised action proposals from spatiotemporal super-voxels}

\date{Received: date}
\author{Mihir Jain
\and Jan van Gemert
\and Herv\'e J\'egou
\and Patrick Bouthemy
\and Cees G.M. Snoek
}

\maketitle

\begin{abstract}
This paper considers the problem of localizing actions in videos as a sequences of bounding boxes. The objective is to generate action proposals that are likely to include the action of interest, ideally achieving high recall with few proposals.
Our contributions are threefold. First, inspired by selective search for object proposals, we introduce an approach to generate action proposals from spatiotemporal super-voxels in an unsupervised manner, we call them \textit{Tubelets}. Second, along with the static features from individual frames our approach advantageously exploits motion. We introduce independent motion evidence as a feature to characterize how the action deviates from the background and explicitly incorporate such motion information in various stages of the proposal generation. Finally, we introduce spatiotemporal refinement of Tubelets, for more precise localization of actions, and pruning to keep the number of Tubelets limited. 

We demonstrate the suitability of our approach by extensive experiments for action proposal quality and action localization on three public datasets: UCF Sports, MSR-II and UCF101. For action proposal quality, our unsupervised proposals beat all other existing approaches on the three datasets. For action localization, we show top performance on both the trimmed videos of UCF Sports and UCF101 as well as the untrimmed videos of MSR-II.
\end{abstract}



\section{Introduction} \label{tubelet_sec:intro}
The goal of this paper is to localize and recognize actions such as `kicking', `hand waving' and `salsa spin' in video content. The recognition of actions has witnessed tremendous progress in recent years thanks to advanced video representations based on motion and appearance \eg \citep{Laptev2005,dollar_vs-pets,wang:ijcv13,Wang2015,SimonyanNIPS14}. However, determining the spatiotemporal extent of an action has appeared considerably more challenging. Early success came from an exhaustive evaluation of possible action locations \eg~\citep{yanke_iccv05,tian_iccv11,Yicong:sdpm}. 
{Such a sliding cuboid is tempting, but owing 
to large number of possible locations demands a relatively simple video representation, \eg~\citep{dalal_hog,hog3D}.}	
Moreover, the rigid cuboid shape does not necessarily capture the versatile nature of actions well. We propose an approach for action localization enabling flexible spatiotemporal subvolumes, while still allowing for modern video representations.

Tran and Yuan pioneered the prediction of flexible spatiotemporal boxes around actions~\citep{Tran:cvpr11,Tran:nips12}. They first obtain for each individual frame the most likely spatial locations containing the action, before determining the best temporal path or \emph{action proposal} through the box search space~\citep{Tran:cvpr11,Tran:nips12}. Surprisingly, the initial spatial classification is frame-based and ignores motion characteristics for action recognition. 
More recently both \cite{gkioxariCVPR15actionTubes} and \cite{Weinzaepfel_iccv15} overcome this limitation by relying on a two-stream convolutional neural network based on appearance and two-frame motion flow.
{While proven effective, these works need to determine the locations in each frame with supervision, and for each action class separately, making them less suited for action localization challenges requiring hundreds of actions.}
Rather than separating the spatial from the temporal analysis and relying on region-level class-specific supervision, we prefer to analyze both spatial and temporal dimensions jointly to obtain action proposals in an unsupervised manner and avoid supervision until classification. Such an approach is easier to scale to hundreds of classes. Moreover, the same set of proposals can be used for applications requiring different encodings or classification schemes. 

\begin{figure*}[t!]
\centering
\includegraphics[width=1\linewidth]{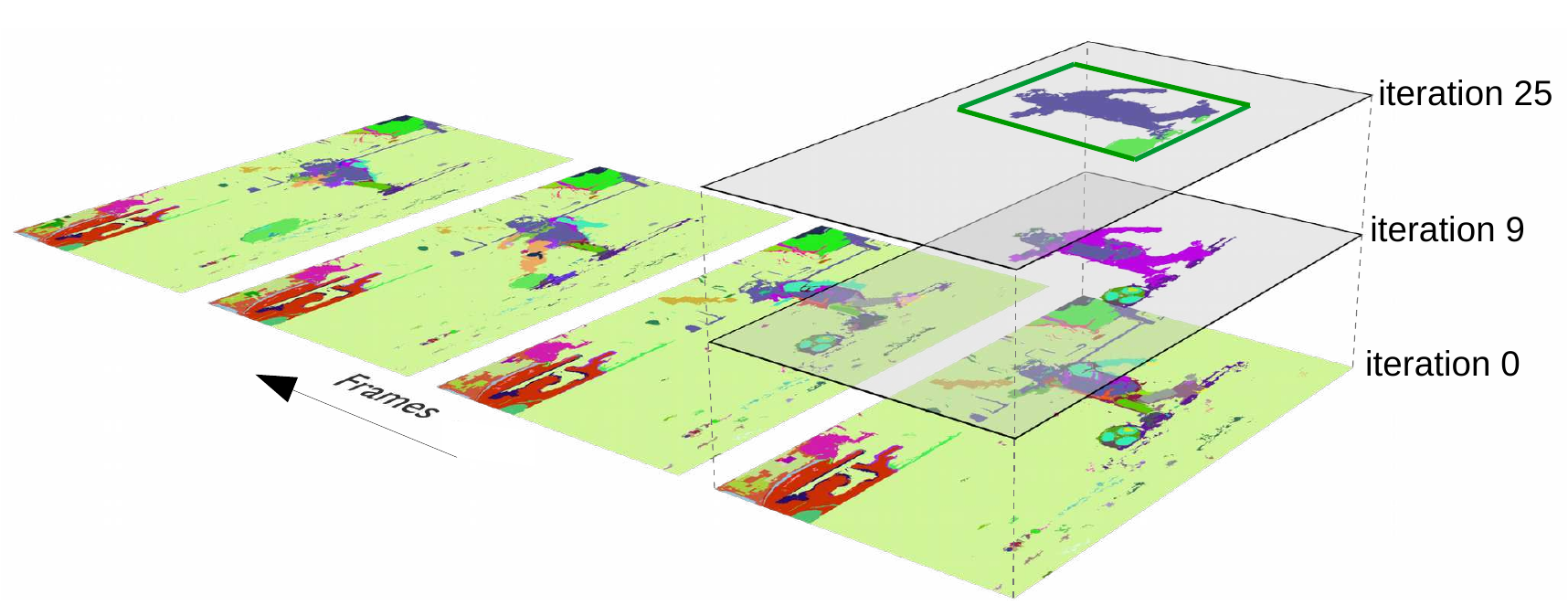}
  \caption{
Overview of unsupervised action proposal from super-voxels: An initial super-voxel segmentation of a video example is shown as a frame sequence in the bottom layer. The proposed grouping (only shown for one frame) iteratively merges the super-voxels that are on the action. One of the better super-voxels after grouping is shown in blue, enclosed by a green box. We refer to the sequence of such bounding boxes over the frames as a Tubelet.} 
\label{tubelet_fig:tubes}
\end{figure*}

{We are inspired by a method for object detection in static images called selective search~\citep{Jasper:selective}.}
The algorithm generates box proposals for possible object locations by hierarchically merging adjacent super-pixels from~\citep{Felzenszwalb2004} 
, based on similarity criteria for color, texture, size and fill. The approach does not require any supervision, making it suited to evaluate many object classes with the same set of proposals. The small set of object proposals is known to result in both high recall and  overlap with the ground-truth \citep{Hosang2016}. Moreover, by separating the localization from the recognition, selective search facilitates modern encodings, such as Fisher vectors of \citep{Sanchez2013} in \citep{SandeCVPR14} and convolutional neural network features in \citep{Girshick2016}. Following the example set by selective search for object detection, we introduce unsupervised spatiotemporal proposals for action localization by relying on video-specific appearance and motion properties derived from super-voxels.

\cite{BroxJMeccv10} realized earlier that temporally consistent segmentations of moving objects in a video can be obtained without supervision. They propose to cluster long term point trajectories and show that these lead to better segmentations than two-frame motion fields. Both \cite{chen2015action} and \cite{Gemert_BMVC15} build on the work of \cite{BroxJMeccv10} and propose action proposals by clever clustering the improved dense trajectories of \cite{wang:imptraj13}. Their approaches are known to be very effective for untrimmed videos where temporal localization is essential. We adopt the use of long term trajectories for temporal refinement and pruning of our action proposals, but we do not restrict ourselves exclusively to improved dense trajectories as representation for action classification.

Our first out of three contributions is to generalize the selective search strategy for unsupervised action proposals in videos. We adopt the general principle designed for static images and repurpose it for video. We consider super-voxels instead of super-pixels to produce spatiotemporal shapes. This directly gives us 2D+t sequences of bounding boxes, without the need to address the problem of linking boxes from one frame to another, as required in other approaches~\citep{Tran:cvpr11,Tran:nips12,gkioxariCVPR15actionTubes,Weinzaepfel_iccv15}. We refer to our action proposal as \emph{Tubelets} in this paper, and summarize their generation in Figure~\ref{tubelet_fig:tubes}.

Our second contribution is explicitly incorporating motion information in various stages 
of the analysis. We introduce \emph{independent motion evidence} as a feature to characterize how the action motion deviates from the background motion. By analogy to image descriptors such as the Fisher vector~\citep{Sanchez2013}, we encode the singularity of the motion in a feature vector associated with each super-voxel. We use the motion as an independent cue to produce super-voxels segmenting the video. In addition, motion is used as a merging criterion in the agglomerative grouping of super-voxels leading to better Tubelets. 

A preliminary version of this article appeared as \cite{Jain:tubelets}. 
{The current version adds as third contribution, the spatiotemporal refinement and pruning of Tubelets. 
The spatiotemporal refinement includes temporal sampling and smoothing the irregular shaped Tubelets. This post-processing considerably improves the performance while keeping the number of proposals manageable.}
Where \cite{chen2015action} and \cite{Gemert_BMVC15} derive their proposals directly and exclusively from the improved dense trajectories, we use the trajectories to refine our unsupervised action proposals from super-voxels. 
{In addition to this technical novelty, the current paper adds: i) detailed experimental evaluation of motion-based segmentation for better proposals, leading to large gains in both proposal quality and action localization, ii) apart from UCF Sports and MSR-II we also consider the much larger UCF101 dataset, iii) revised experiments for all three datasets considering both the quality of the proposal as well as their suitability for action localization 
using modern video representations \citep{Sanchez2013,googlenet}, and iv) a new related work section, which will be discussed next.}

\section{Related work}
\label{tubelet_sec:related}

\rowcolors{5}{white}{gray!10}

\begin{table*}
\centering
\scalebox{0.95}{
\begin{tabular}{p{1.5cm}p{2.5cm}p{3.4cm}p{2.71cm}p{3.02cm}p{2.6cm}} \toprule
 Representation & \multicolumn{5}{c}{Approach} \\ \cmidrule(lr){1-1} \cmidrule(lr){2-6}
 & \multicolumn{2}{c}{2D Detect and track}  & \multicolumn{3}{c}{3D spatio-temporal volume} \\ \cmidrule(lr){2-3} \cmidrule(lr){4-6}
 & \multicolumn{1}{c}{Human detector}  & \multicolumn{1}{c}{Generic detector} & \multicolumn{1}{c}{Cuboid} & \multicolumn{1}{c}{Trajectory} & \multicolumn{1}{c}{Voxels} \\  \cmidrule(lr){2-2} \cmidrule(lr){3-3} \cmidrule(lr){4-4} \cmidrule(lr){5-5} \cmidrule(lr){6-6}
-  & 
& 
\cite{marianICCV2015unsupervisedTube}
&
\cite{chenCVPR14actionness}
& 
& 
\cite{oneataECCV14spatemprops} 
\\
Part-based & 
\cite{tian_iccv11} \newline
\cite{Wang_dynamicPoslets} 
& 
&
\cite{Yicong:sdpm}
& 
\cite{Raptis:cvpr12}
& 
\\
Cube & 
\cite{klaser2012human}
& 
\cite{Tran:nips12}
&
\cite{yanke_iccv05} \newline
\cite{yuan_cvpr09} \newline
\cite{Cao:cvpr10} \newline
\cite{derpanisPAMI2013actionSpotting}
& 
& 
\\
BoW & 
\cite{ma2013ICCVspacetimesegments} 
& 
\cite{Tran:cvpr11} 
&
& 
\cite{mosabbebACCV2014weakly} \newline
\cite{chen2015action} 
& 
\cite{Jain:tubelets} \newline
\cite{soomro2015ICCVcontextWalk} \newline
\textbf{This paper}
\\
Fisher & 
\cite{yuCVPR15fasttubes} 
& 
&
& 
\cite{Gemert_BMVC15}
& 
\textbf{This paper} 
\\
CNN & 
& 
\cite{gkioxariCVPR15actionTubes}
&
& 
& 
\cite{jain2015objects2action} \newline
\textbf{This paper}
\\
CNN+Cube & 
& 
\cite{Weinzaepfel_iccv15}
&
& 
& 
\\
CNN+BoW & 
& 
&
& 
& 
\cite{Jain:15kobjects}
\\
CNN+Fisher & 
& 
&
& 
& 
\textbf{This paper}
\\
\bottomrule
\end{tabular}
}
\caption{Related work linking the action representation with approaches in action localization. Our work does not treat a video as a collection of 2D frames. Instead, we take a holistic spatiotemporal approach by aggregating 3D voxels. From these voxels we build Tubelets on which we evaluate several state-of-the-art action representations.}
\label{tab:relatedWork}
\end{table*}

\rowcolors{1}{white}{white}

We discuss action recognition and action localization. In Table~\ref{tab:relatedWork} we link action recognition representations with action localization methods and use it to structure our discussion of related work.

\subsection{Action recognition}

\textbf{Part-based} Action recognition by parts typically exploits the human actor.
Correctly recognizing the human pose improves performance~\cite{Jhuang_2013_ICCVtowardsUnderstandAR}. A detailed pose model can make fine-grained distinctions between nearly similar actions~\cite{Cheron_2015_ICCVposeCNN}. Pose can be modeled with poselets~\cite{majiCVPR2011actionPose} or as a flexible constellation of parts in a CRF~\cite{wangPAMI2011hiddenPart}. For action recognition in still images where motion is not available the human pose can play a role~\cite{delaitreBMVC2010bowAndParts} as modeled in a part-based latent SVM~\citep{Felzenszwalb:DPM}. In our work we make no explicit assumptions on the pose, and use generic local video features.

\textbf{Cube} Local video features are typically represented by a 3D cube. The seminal work of~\citep{Laptev2005} on Spatio-Temporal Interest Points (STIPs) detects points that are salient in appearance and motion and then uses a cube of Gaussian derivative filter responses to represent the interest points. An alternative representation is HOG3D~\cite{hog3D} which extends the 2D Histogram of Oriented Gradients (HOG) of~\cite{dalal_hog} to 3D. Instead of using sparse salient points, the work of \cite{dollar_vs-pets} shows that using denser sampling improves results. Replacing dense points with dense trajectories~\citep{Wang2015} and flexible track-aligned feature cubes with motion boundary features yields excellent performance. The improved trajectories take into account the camera motion compensation, which is shown to be critical in action recognition~\citep{jain2016improved,Piriou_ieee-tip2006,wang:imptraj13}. In our work we build on these dense trajectories as well.

\textbf{Bag of Words} To arrive at a global representation over all local descriptors, BoW represents a cube descriptor by a prototype. The frequency of the prototypes aggregated in a histogram is a global video representation. The BoW representation is simple and offers good results~\citep{evertsTIP14evaluation,densetrack11}. We consider BoW as one of our representations for action localization as well.

\textbf{Fisher Vector} Where BoW records prototype frequency counts, the Fisher vector~\citep{Sanchez2013} and the VLAD \citep{JPDSPS12} model the relation between local descriptors and prototypes in the feature space of the descriptor. This more sophisticated variant of BoW outperforms BoW~\citep{Jain:wflow,Oneata:iccv13,oneata:cvpr_14}. Because of the good performance we also consider the Fisher vector as a representation.

\textbf{CNNs} Deep learning on visual data with CNNs (Convolutional Neural Networks) has revolutionized static image recognition~\cite{Krizhevsky_imagenetclassification}. For action recognition in videos, the work of \cite{SimonyanNIPS14} separate video in two channels: a network on static RGB and a network on hand-crafted optical flow. In~\cite{wangCVPR15tdd} CNN features are used as a local feature in dense trajectories using a Fisher vector. Long term motion can be modeled by recurrent networks~\cite{yueCVPR15beyondShortSnippets}. The distinction between motion and static objects is analyzed in~\cite{Jain:15kobjects} and extended by~\cite{jain2015objects2action} for action recognition without using any video training data. Instead of separating static and motion, 3D convolutional networks combine both~\cite{tranICCV2015c3d}. Due to excellent performance we also adopt CNN features as a representation for action localization.

\subsection{Action localization}

\textbf{2D Human detector} Spatiotemporal action localization can be realized by running a human detector on each frame and tracking the detections. In~\cite{klaser2012human} a sliding window upper-body HOG detector per frame is tracked by optical flow feature points for spatial localization. Temporal localization is achieved with a sliding window on track-aligned HOG3D features.  HOG3D features are also used in~\cite{tian_iccv11} albeit in BoW, where the 2D person detector is treated as a latent variable and an undirected relational graph inspired by a latent SVM is used for classification. Similarly, the human pose is used by~\cite{Wang_dynamicPoslets} in a relational dynamic poselet model using cuboids to model a mixture of parts. In~\cite{ma2013ICCVspacetimesegments} dynamic action parts are extended by incorporating static parts using 2D segments. Segments are grouped to tracks and represented in a hierarchical variant of BoW. In our work we do not make the assumption that an action has to be performed by a human. Our method is equally applicable to actions by groups, animals, or vehicles.

\textbf{2D generic detector} By replacing the human detector with a generic detector the types of actions can be extended beyond a human actor. This can be done by finding the best path trough fixed positions in a frame using HOG/HOF directly~\citep{Tran:nips12} or through BoW~\citep{Tran:cvpr11}. Instead of fixed positions, \cite{gkioxariCVPR15actionTubes} classify object proposals with a two-stream CNN and track overlapping proposals with a high classification score. The work of~\cite{Weinzaepfel_iccv15} uses a similar two-stream CNN approach, adding a HOG/HOF/MBH-like cube descriptor at the track level and add temporal localization with a sliding window. The need for strong supervision is removed by~\cite{marianICCV2015unsupervisedTube} where generic CNN feature are linked through dense trajectory tracks to yield action proposals that could be used for action localization. Similarly, our work requires no supervision for obtaining action proposals, and we experimentally show that these proposals give good results. In addition, we do not first treat a video as a collection of static frames where temporal relations are added as an separate second step. Instead, we respect the 3D spatiotemporal nature of video from the very beginning.

\textbf{3D Trajectory} The strength of 3D dense trajectories~\cite{Wang2015} for action recognition spilled over to action localization. In~\cite{Raptis:cvpr12} mid-level clusters of trajectories are grouped and matched with a graphical model. The work of~\cite{mosabbebACCV2014weakly} groups trajectories to parts which are used in a BoW in an unsupervised manner using low-rank matrix completion and subspace clustering. Similarly, BoW on space-time graph clusters is used by~\cite{chen2015action}  and a Fisher vector on trajectories is used on hierarchical clusters in~\cite{Gemert_BMVC15} for action localization. These methods specifically target the strength of dense trajectories. Instead, our approach does not commit itself to a single representation.

\textbf{3D Cuboid} The 3D nature of video is respected by building on space-time cuboids for action localization. Such cuboids are a natural extension of 2D patches to 3D. \cite{yanke_iccv05} offer a 3D extension of the seminal face detector of~\cite{violajones_ijcv} using 3D cuboids with optical flow features.  The work of~\cite{yuan_cvpr09} and \cite{Cao:cvpr10} exploit the efficient branch and bound method~\citep{Lampert:ESS} in 3D. In~\cite{Yicong:sdpm} the deformable part-based model~\citep{Felzenszwalb:DPM} is generalized to 3D, an efficient sliding window approach in 3D is proposed by~\cite{derpanisPAMI2013actionSpotting} and ordinal regression~\citep{kimECCV2010ordinalRegression} is extended by~\cite{chenCVPR14actionness}. Instead of using cuboids, which are rigid in time and space, we choose a more delicate approach using 3D voxels.

\textbf{3D Voxels} As a 3D generalization of 2D image segmentation the voxels from video segmentation methods~\citep{xuCVPR12SuperVoxelEvaluation} offer flexible and fine-grained tools for action proposals. In extension of~\cite{manenICCV13objectProposals}, the work of~\cite{oneataECCV14spatemprops} groups voxels together for action proposals using minimal training. Such action proposals could be used for action localization. This is done by~\cite{soomro2015ICCVcontextWalk} who use a supervised CRF to model foreground-background relationships for proposals and action localization. Instead, our proposal method is unsupervised and thus class agnostic. This is beneficial as this makes our algorithm independent on the number of action classes. This paper is an extension of~\cite{Jain:tubelets}, where 3D voxels are grouped to proposals based on features such as color, texture and motion. The proposals have successfully been used for action localization using objects~\cite{Jain:15kobjects} and in a zero-shot setting~\cite{jain2015objects2action}. We will discuss the mechanics of our unsupervised action proposals next.

\def\smotion{s_\mathrm{M}}
\def\scolor{s_\mathrm{C}}
\def\stexture{s_\mathrm{T}}
\def\sfill{s_\mathrm{F}}
\def\ssize{s_\Gamma}

\def\hmotion{h_\mathrm{M}}
\def\hcolor{h_\mathrm{C}}
\def\htexture{h_\mathrm{T}}
\def\h{h}
\def\s{s}

\begin{figure*}[th]
\centering
 \includegraphics[width=0.99\linewidth]{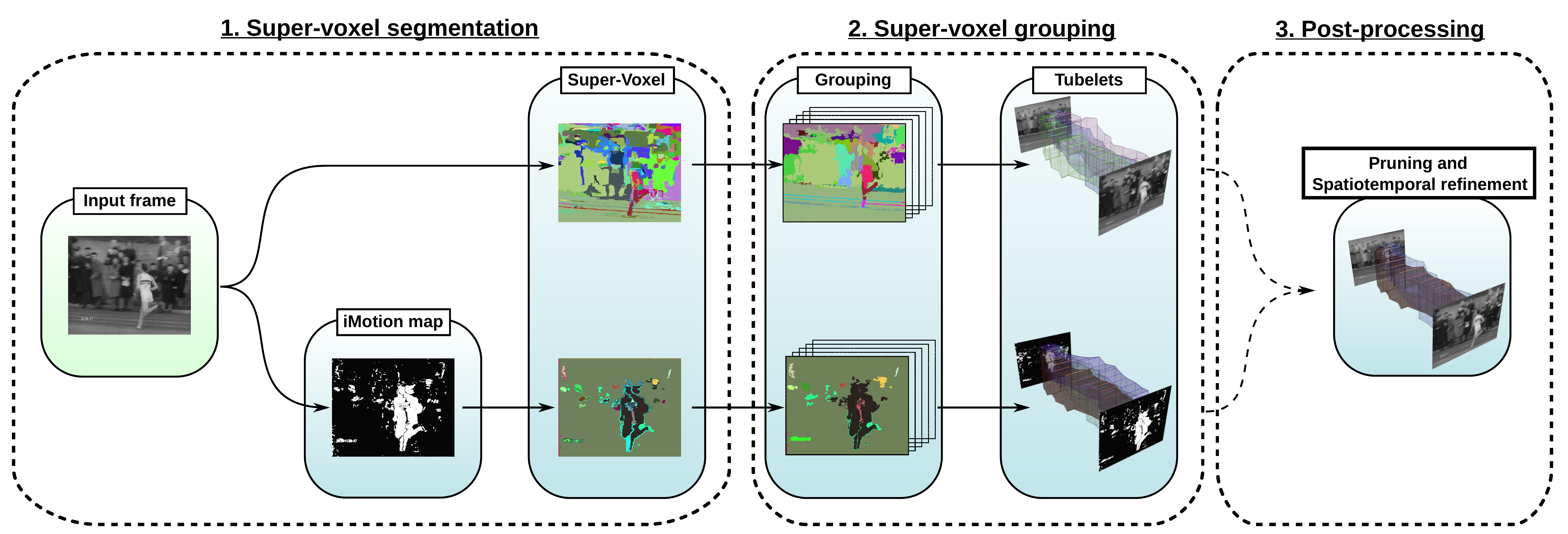}
 \caption{Tubelet generation: In the first stage a video is segmented into super-voxels. In addition to segmenting video frames, we also segment their \imemotion maps to also include motion information in the \emph{super-voxel segmentation} stage. In the second stage of \emph{super-voxel grouping}, super-voxels are iteratively merged using several \emph{grouping functions} each of them leading to a set of action proposals. These sets are again grouped by union into a set of Tubelets. The final stage is post-processing that includes pruning and spatiotemporal-refinement of action proposals.}
\label{tubelet_pipeline}
\end{figure*}

\section{Unsupervised action proposals: Tubelets} \label{tubelet_sec:tubes}
In this section we present our approach to obtain action proposals from video in an unsupervised manner, we call the spatiotemporal proposals \emph{Tubelets}. The three stages of the Tubelet generation process are shown in Figure~\ref{tubelet_pipeline}.  
We first introduce in Subsection~\ref{tubelet_sec:ime} our motion model based on evidence of independent motion. This motion cue is used in the first two stages of the process. In Subsection~\ref{tubelet_sec:ime_seg}, we discuss the first stage, \emph{super-voxel segmentation}, to generate an initial set of super-voxels from video. For this we rely on an off-the-shelf video segmentation as well as our proposed independent motion evidence. In Subsection~\ref{tubelet_sec:merge} we detail the second stage of \emph{super-voxel grouping}, where we iteratively group the two most similar super-voxels into a new one. 
The similarity score is computed using multiple \emph{grouping functions}, each leading to a set of super-voxels. A super-voxel is tightly bounded by a rectangle in each frame it appears. The temporal sequence of bounding boxes forms our action proposal, a Tubelet. In Subsection~\ref{tubelet_sec:tempo_samp}, we introduce spatiotemporal refinement and pruning of Tubelets. This enhances the proposal quality, especially for temporal localization, while at the same time keeping the number of proposals feasible to use computationally expensive features and memory demanding encodings for action localization.

\subsection{Evidence of independent motion} \label{tubelet_sec:ime}
Since we are concerned with action localization, we need to aggregate super-voxels corresponding to the action of interest. Most of the points in such super-voxels would deviate from the background motion caused by moving camera and usually assumed to be dominant motion.
In other words, the regions corresponding to independently moving objects do not, usually, conform with the dominant motion in the frame. 
The dominant frame motion can be represented by a 2D parametric motion model. Typically, an affine motion model of parameters $\theta=(a_i)$, $i=1...6$, or a quadratic (perspective) model with 8 parameters can be used, depending on the type of camera motion and the scene layout likely to occur:
\begin{align*}
w_{\theta}(p)            = & (a_1 + a_2 x + a_3 y,a_4 + a_5 x + a_6 y) \\
\mbox{or~} w_{\theta}(p) = & (a_1 + a_2 x + a_3 y + a_7 x^2 + a_8 xy,\\
                           & a_4 + a_5 x + a_6 y + a_7 xy + a_8 y^2),
\end{align*}
where $w_{\theta}(p)$ is the velocity vector supplied by the motion model at point $p=(x,y)$ in the image domain $\Omega$. In this paper, we use the affine motion model for all the experiments.

We formulate the evidence that a point $p \in \Omega$ undergoes an independent motion (\ie, an action related motion) at time step $t$. Let us introduce the displaced frame difference at point $p$ and at time step $t$ for the motion model of parameter $\theta_t$: $r_{\theta_t}(p,t) = I(p+w_{\theta_t}(p),t+1) - I(p,t)$. Here, $r_{\theta_t}(p,t)$ will be close to $0$ if point $p$ only undergoes the background motion due to camera motion.
At every time step $t$, the global parametric motion model can be estimated with a robust penalty function as 
\begin{equation}
\label{motion2d}
\hat{\theta_t} = \arg \min_{\theta_t} \sum_{p\in\Omega}\rho(r_{\theta_t}(p,t)),
\end{equation}
where $\rho$ is the robust function. To solve (\ref{motion2d}), we use the publicly available Motion2D software by ~\citep{Odobez95robustmultiresolution}, where $\rho(.)$ is defined as the Tukey function. $\rho(r_{\theta_t})$ produces a maximum likelihood type estimate: the so-called M-estimate~\citep{Huber81}. Indeed, if we write $\rho(r_{\theta_t}) = - \log f(r_{\theta_t})$ for a given function~$f$, $\rho(r_{\theta_t})$ supplies the usual maximum likelihood estimate. Since we are looking for action related moving points in the image, we want to measure \emph{the deviation} to the global (background) motion. This is in spirit of the Fisher vectors 
by ~\citep{PD07}, where the deviation of local descriptors from a background Gaussian mixture model is encoded to produce an image representation. 

Let us consider the derivative of the robust function $\rho(.)$. It is usually denoted as $\psi(.)$ and corresponds to the influence function~\citep{Huber81}. More precisely, the ratio $\psi(r_{\theta_t})/r_{\theta_t}$ accounts for the influence of the residual $r_{\theta_t}$ in the robust estimation of the model parameters. The higher the influence, the more likely the point conforms to the global motion. Conversely, the lower the influence, the less likely the point approves to the global motion. This leads us to define the \emph{independent motion evidence} as:
\begin{equation}
\xi(p,t) = 1 - \varpi(p,t),
\end{equation}
where $\varpi(p,t)$ is the ratio $\frac{\psi(r_{\hat{\theta_t}}(p,t))}{r_{\hat{\theta_t}}(p,t)}$ normalized within $[0,1]$. 

\begin{figure*}[th!]
  \subfigure[Video frames] {\label{tubelet_fig:vid_frames}
  \includegraphics[height=4cm]{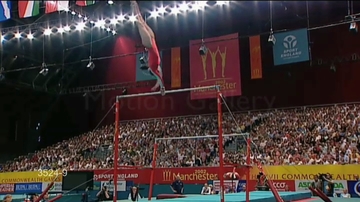}
  \includegraphics[height=4cm]{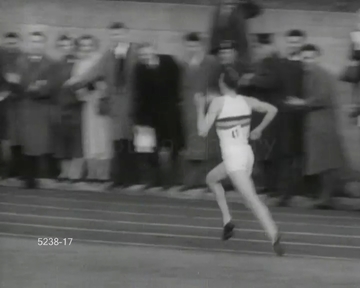}       
  \includegraphics[height=4cm]{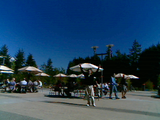} }

  \subfigure[Independent motion in frames] {\label{tubelet_fig:ime_frames}
  \includegraphics[height=4cm]{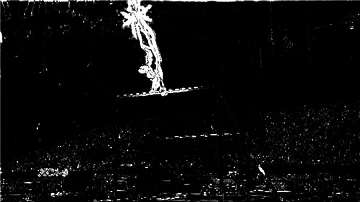}
  \includegraphics[height=4cm]{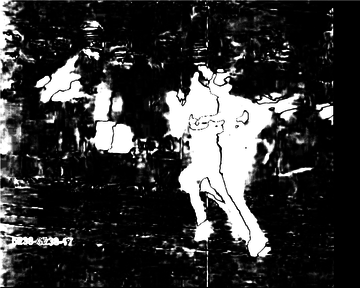}
  \includegraphics[height=4cm]{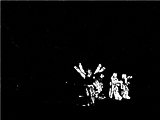} }

  \subfigure[\imemotion maps] {\label{tubelet_fig:ime_maps}
  \includegraphics[height=4cm]{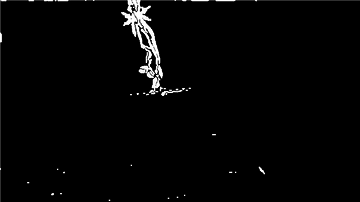}
  \includegraphics[height=4cm]{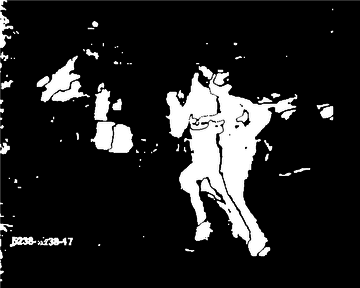}
  \includegraphics[height=4cm]{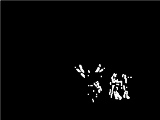} }

  \subfigure[Graph-based segmentation of \imemotion maps (each color represents a super-voxel)] {\label{tubelet_fig:ime_seg}
  \includegraphics[height=4cm]{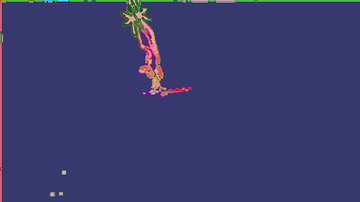}
  \includegraphics[height=4cm]{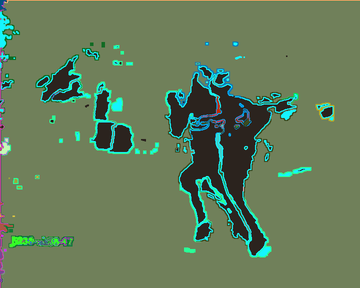}
  \includegraphics[height=4cm]{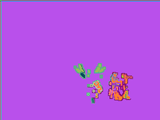} }
\caption{\imemotion maps for segmentation: Top two rows show the original frames and their independent motion. The \imemotion maps obtained after applying morphological operations are shown in the third row. The bottom row shows the result of applying graph-based video segmentation on \imemotion maps. The process is illustrated for three example video clips for actions `Swing-Bench', `Running' and `Hand Waving' respectively. In spite of clutter and illumination variations the \imemotion map successfully highlights the action.} \label{tubelet_fig:ime}
\end{figure*}

\begin{figure*}
\centering
 \includegraphics[width=0.99\linewidth]{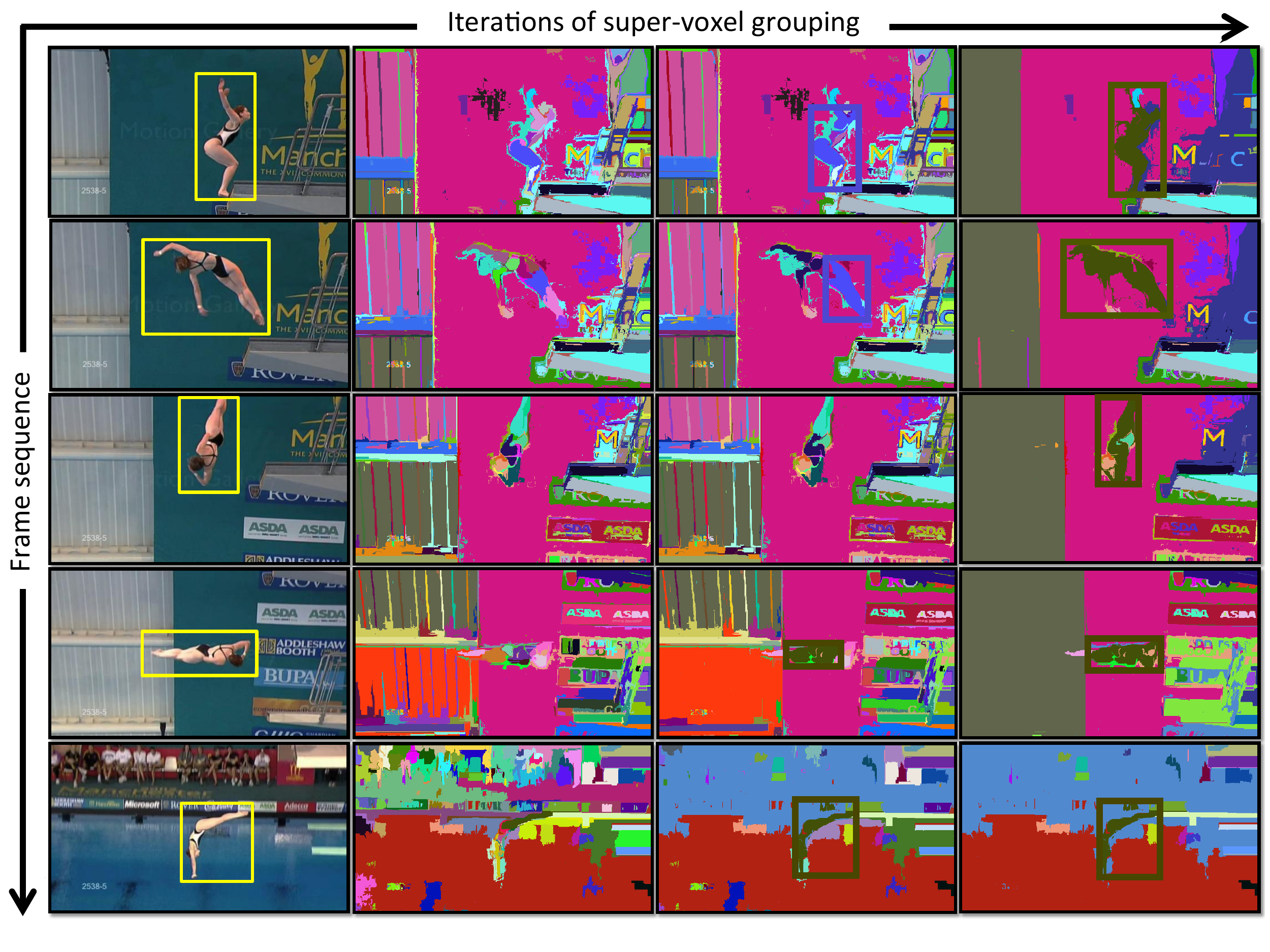}
  \caption{Illustration of hierarchical grouping of super-voxels into Tubelets. Left column: A sampled sequence of frames (1\textsuperscript{st}, 15\textsuperscript{th}, 25\textsuperscript{th}, 35\textsuperscript{th}, 50\textsuperscript{th}) associated with the action `Diving'. The yellow bounding boxes represent the ground-truth sequence. Column~2: the initial video segmentation used as input to our method. The last two columns show the two junctures of the iterative grouping algorithm. A Tubelet close to the action is also represented by bounding boxes in these two columns. Observe how close it is to the ground-truth in the last column despite the varying aspect ratios in different frames.}
\label{tubelet_fig:3dSS}
\end{figure*}

\subsection{Super-voxel segmentation} \label{tubelet_sec:ime_seg}
To generate an initial set of super-voxels, we rely on a third-party graph-based video segmentation by~\citep{xuCVPR12SuperVoxelEvaluation}. We choose their graph-based segmentation over other methods in~\citep{xuCVPR12SuperVoxelEvaluation} because it is more efficient w.r.t. time and memory. The graph-based segmentation is about 13 times faster than the slightly more accurate hierarchical version \citep{xuCVPR12SuperVoxelEvaluation}.

\paragraph{Independent motion.} 
{As an alternative to the off-the-shelf video segmentations, each video frame is represented with the corresponding map, $\xi(t)$, of independent motion of pixels. This encodes motion information in the segmentation. We show video frames and their $\xi(t)$ maps in Figure~\ref{tubelet_fig:vid_frames} and~\ref{tubelet_fig:ime_frames}.
We post-process the independent motion or $\xi(t)$ maps by applying a morphological closing operation (dilation followed by erosion) to obtain denoised maps, which we refer to as \imemotion maps, displayed in Figure~\ref{tubelet_fig:ime_maps}.} 
Applying the graph-based video segmentation of~\citep{xuCVPR12SuperVoxelEvaluation} on sequences of these denoised maps partitions the video into super-voxels with independent motion. 
Three examples of results obtained this way are shown in Figure~\ref{tubelet_fig:ime_seg}. The first column shows a frame from action `Swing-Bench', where the action of interest is highlighted by \imemotion map itself and then clearly delineated by segmenting the maps. Second column shows an example from action `Running'. Here the segmentation does not give an ideal set of initial super-voxels but the \imemotion map has useful information to be exploited by our motion feature based merging criterion (described in Subsection~\ref{tubelet_sec:merge}).  An example of `Hand Waving' is shown in the last column. 
The resulting super-voxels are more adapted and aligned to the action sequences. This alternative for initial segmentation is also more efficient, about 4 times faster than graph-based segmentation on the original video and produces 8 times fewer super-voxels.
{Unlike graph-based video segmentation on original frames this alternate set of initial super-voxels exploits motion information. The two are complementary and together lead to much better proposal quality as shown later in our experiments.}

\begin{figure*}[t]
\centering
\includegraphics[height=2.72cm]{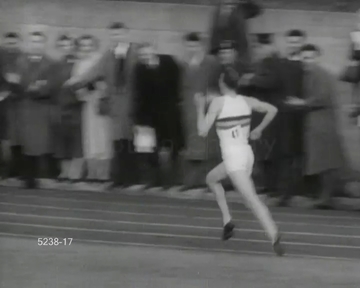} 			
\includegraphics[height=2.72cm]{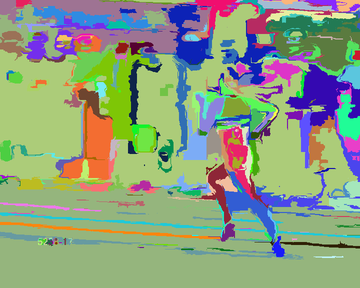} 	 
\includegraphics[height=2.72cm]{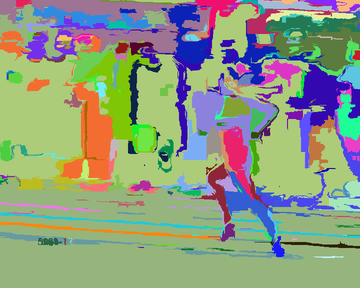} 	
\includegraphics[height=2.72cm]{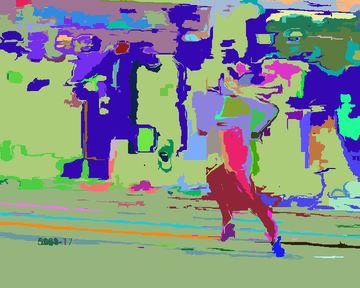} 	  
\includegraphics[height=2.72cm]{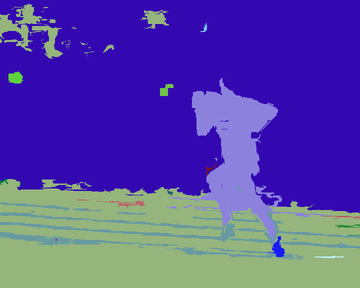} 
\caption{Example for the action `Running': The first two images depict a video frame and the initial super-voxel segmentation used as input of our approach. The next three images represent the segmentation after a varying number of merge operations.}
\label{tubelet_fig:run}
\end{figure*}

\subsection{Super-voxel grouping} \label{tubelet_sec:merge} 
{Having defined our ways to segment a video sequence into super-voxels, we are now ready to present our method for grouping super-voxels into Tubelets. The grouping is done in two steps. In the first step, initial super-voxels are grouped iteratively to create new super-voxels. A grouping function computes the similarity between any two super-voxels and the successive groupings of the most similar pairs lead to a new set of super-voxels. Each grouping function leads to a set of super-voxel. In the second step, the super-voxel sets produced by multiple grouping functions are again grouped by union. This united set of super-voxels is then enclosed by boxes in each frame to yield the Tubelets.}

\paragraph{Iterative grouping.} 
We iteratively group super-voxels in an agglomerative manner.  Starting from the initial set of super-voxels, we hierarchically group them until the video becomes a single super-voxel. At each iteration, a new super-voxel is produced from two super-voxels, which are then not considered any more in subsequent iterations. This iterative merging algorithm is inspired by the selective search method proposed for localization in images by~\citep{Jasper:selective}. 

Formally, we produce a hierarchy of super-voxels that are represented as a tree: The leaves correspond to the $n$ initial super-voxels while the internal nodes are produced by the merge operations. The root node is the whole video and the corresponding super-voxel is produced in the last iteration. Since this hierarchy of super-voxels is organized as a binary tree, it is straightforward to show that $n-1$ additional super-voxels are produced by the algorithm. Out of these $n-1$ super-voxels, those which are very small or contain no motion at all are discarded at this point.
This usually leaves much fewer number of super-voxels depending upon the grouping function used.

\paragraph{Grouping function.} 
{For selection of the two super-voxels to be grouped, we rely on similarities computed between all the neighboring super-voxels that are still active.} We employ five complementary similarity measures in our grouping functions to compare super-voxels, in order to decide which should be merged. They are fast to compute. Four of these measures are adapted from selective search in image: The measures based on Color, Texture, Size and Fill were computed for super-pixels~\citep{Jasper:selective}. We revise them for super-voxels. As our objective is not to segment the objects but to delineate the actions or actors, we additionally employ a motion-based similarity measure based on our independent motion evidence to characterize a super-voxel. The grouping function is defined as any one of the similarity measures or sum of multiple of them. Next we present the five similarity measures for super-voxels: \emph{motion}, \emph{color}, \emph{texture}, \emph{size} and \emph{fill}.

\begin{figure*}[t]
\centering
 \includegraphics[width=0.99\linewidth]{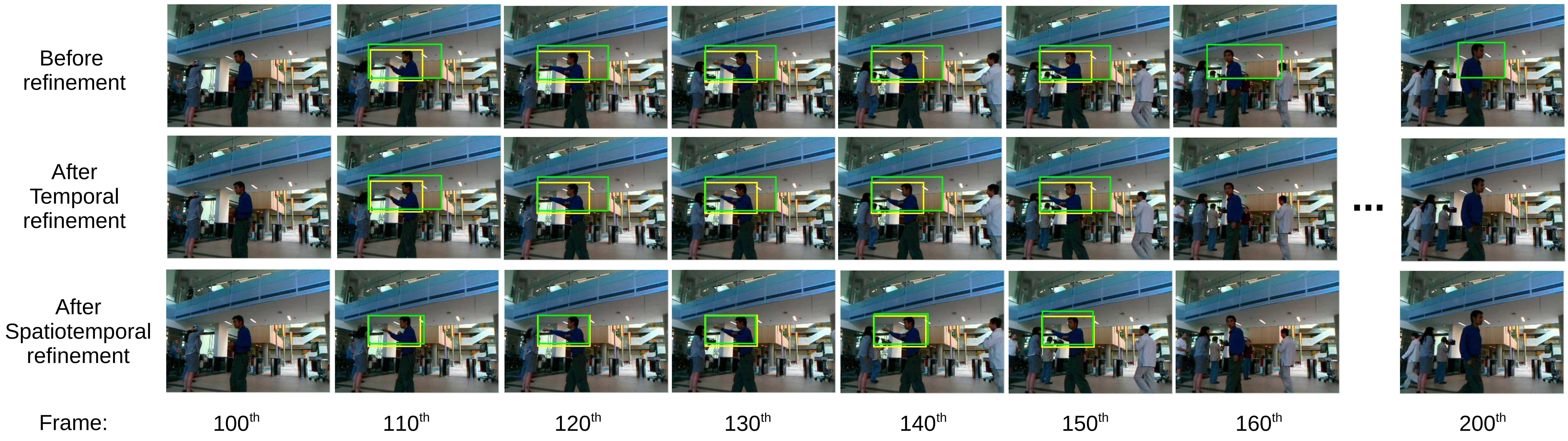}
 \caption{Impact of spatiotemporal-refinement of Tubelets: The first row shows an untrimmed video of about 900 frames. The ground-truth action is an instance of `Boxing' from frame 108 to frame 151, as bounded by the yellow boxes. The green boxes in the top row show one of the best Tubelet action proposals obtained for this video. While it aligns well with the ground-truth spatially, it fails temporally as it continues beyond 200 frames. With temporal refinement in the second row, we are able to sample a sub-sequence that localizes the action temporally well also. Third row shows further improvement by spatial refinement.}
\label{tubelet_refinement}
\end{figure*}

\paragraph{Similarity by motion ($\smotion$):} 
We define a motion representation of super-voxels from \imemotion maps capturing the relevant motion information. This motion representation is also efficient to compute. 
{We consider the binarized version of \imemotion maps obtained by setting all non-zero values to $1$.}
{At every pixel $p$, we count the number of pixels $q$  (including $p$) in its $3D$ neighborhood that are set to $1$ (\ie pixels likely to be related to actions).} In a subvolume of $5\times5\times3$ pixels, this count value ranges from 0 to 75.  A motion histogram of these values, denoted by ${\hmotion}_i$, is computed over the super-voxel $r_i$. Intuitively, this histogram captures both the density and the compactness of a given region with respect to the number of points belonging to independently moving objects. 

Now, two super-voxels, $r_i$ and $r_j$, represented by motion histograms are compared as follows. The motion histograms are first $\ell_1$-normalized and then compared with histogram intersection, $\s = \delta_1({\hmotion}_i,{\hmotion}_j)$. The histograms are efficiently propagated through the hierarchy of super-voxels. Denoting with $r_t=r_i \cup r_j$ the super-voxel obtained by merging the super-voxels $r_i$ and $r_j$, we have:
\begin{equation} 
{\hmotion}_k  = \frac{\size(r_i) \times {\hmotion}_i + \size(r_j) \times {\hmotion}_j}{\size(r_i) + \size(r_j)}
\label{eq:propag}
\end{equation}
where $\size(r)$ denotes the number of pixels in super-voxel $r$. The size of the new super-voxel $r_t$ is $\size(r_k) = \size(r_i) + \size(r_j)$.

\paragraph{Similarity by color ($\scolor$) and texture ($\stexture$).} 
In addition to motion, we also consider similarity based on color and texture. Both $\hcolor$ and $\htexture$ are identical to the histograms considered for selective search in images~\citep{Jasper:selective}, be it that we compute them on super-voxels rather than super-pixels. The histograms are computed from color and intensity gradient for each given super-voxel:
\begin{itemize}
\item The color histogram $\hcolor$ captures the HSV components of the pixels included in a super-voxel;
\item $\htexture$ encodes the texture or gradient information of a given super-voxel.
\end{itemize}
The method of similarity computation and the process of merging for color and texture is the same as for motion: Describe each super-voxel with a histogram and compare the two by histogram intersection.

\paragraph{Similarity by size ($\ssize$) and fill ($\sfill$).} The similarity $\ssize(r_i,r_j)$ aims at merging smaller super-voxels first:
\begin{equation} 
\ssize(r_i,r_j) = 1 - \frac{\size(r_i) + \size(r_j)}{\size(\textrm{video})}
\label{eq:size}
\end{equation}
where $\size(\textrm{video})$ is the size of the video (in pixels). This tends to produce super-voxels, and therefore Tubelets, of varying sizes in all parts of the video (recall that we only merge contiguous super-voxels).

The last similarity measure $\sfill$ measures how well super-voxels $r_i$ and $r_j$ fit into each other. We define $B_{i,j}$ to be the tight bounding cuboid enveloping $r_i$ and $r_j$. The similarity is given by:
\begin{equation} 
\sfill(r_i,r_j) = \frac{\size(r_i) + \size(r_j)}{\size(B_{i,j})}.   
\label{eq:fill}
\end{equation}

After each merge, we compute the new similarities between the resulting super-voxel and its neighbors. As illustrated in the following two figures. Figure~\ref{tubelet_fig:3dSS} illustrates the method on a sample video. Each color represents a super-voxel and after every iteration a new super-voxel is added and two are removed. After $1,000$ iterations, observe that two Tubelets (blue and dark green) emerge around the action of interest in the beginning and the end of the video, respectively. At iteration 1,720, the two corresponding super-voxels are merged. The novel Tubelet (dark green) resembles the yellow ground-truth sequence of bounding-boxes. This exhibits the ability of our method to group super-voxels both spatially \emph{and} temporally. Also importantly, it shows the capability to sample an action proposal with boxes having very different aspect ratios. This is unlikely to be coped by sliding-subvolumes or even approaches based on efficient sub-window search. Figure~\ref{tubelet_fig:run} depicts another example, with a single frame considered at different stages of the algorithm. Here the initial super-voxels (second image in first row) are spatially more decomposed because the background is cluttered both in appearance and in motion (spectators cheering). Even in such a challenging case our method is able to group the super-voxels related to the action of interest.

\subsection{Pruning and spatiotemporal refinement of Tubelets} \label{tubelet_sec:tempo_samp}

\paragraph{Pruning proposals}. 
We apply two types of pruning to reduce the number of proposals leading to a more compact set of Tubelet action proposals with minimal impact on the recall.

{\emph{Motion pruning}: The first type of pruning is based on the amount of motion.
Long videos that have much background clutter due to unrelated actors/objects, usually result in many irrelevant Tubelet proposals. We filter them based on their motion content, which we quantify by the number of motion trajectories~\citep{wang:imptraj13}. For each video, we rank the Tubelet proposals based on the number of trajectories, keep the top $P$ proposals and the top ten percent of the rest. This is to ensure that at least a minimal number of proposals are retained from each video.}

{\emph{Overlap pruning}: The second type of pruning is based on mutual overlaps of the action proposals. Many proposals have very high alignment or overlaps between them, all practically representing the same part of the video. To eliminate such redundant proposals we keep only one in a set of many highly overlapping ones. 
It is particularly useful when there is a large number of action proposals per video.}

\begin{figure*}[t]
\centering{
  \renewcommand{\tabcolsep}{4pt}
  \begin{tabular}{cccccc}
  \rotatebox{90}{\bf ~~~~UCF Sports} &
    \includegraphics[width=0.18\linewidth,height=2.4cm]{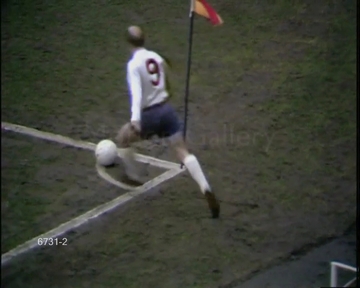} &
    \includegraphics[width=0.18\linewidth,height=2.4cm]{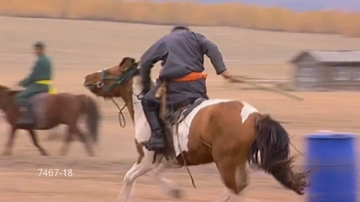} &
    \includegraphics[width=0.18\linewidth,height=2.4cm]{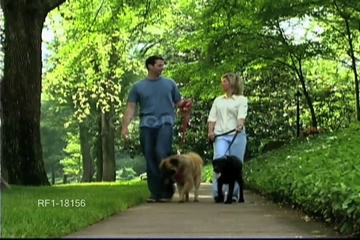}   &
    \includegraphics[width=0.18\linewidth,height=2.4cm]{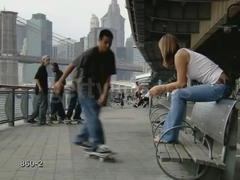}   &
    \includegraphics[width=0.18\linewidth,height=2.4cm]{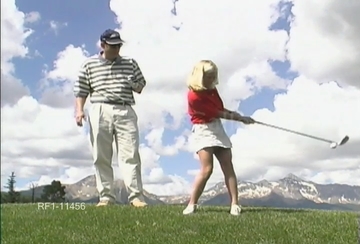}  \\ 
   & \emph{Kicking}  & \emph{Riding-horse} & \emph{Walking}  &  \emph{Skateboarding}  &  \emph{Golf-swinging} \\ \\
  \rotatebox{90}{\bf ~~~~~~~~MSR-II} &
    \includegraphics[width=0.18\linewidth,height=2.4cm]{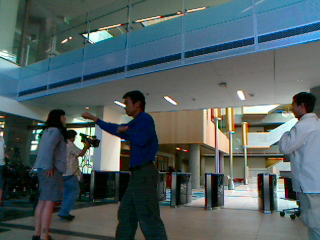} &
    \includegraphics[width=0.18\linewidth,height=2.4cm]{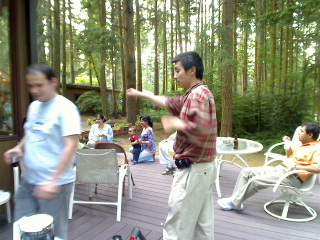} &
    \includegraphics[width=0.18\linewidth,height=2.4cm]{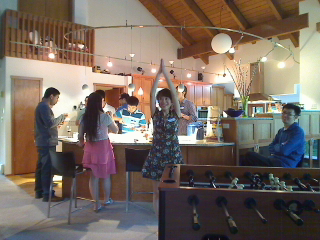} &
    \includegraphics[width=0.18\linewidth,height=2.4cm]{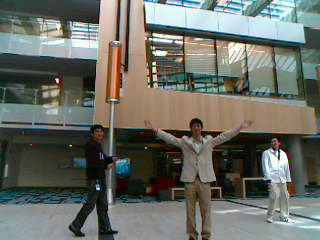} &
    \includegraphics[width=0.18\linewidth,height=2.4cm]{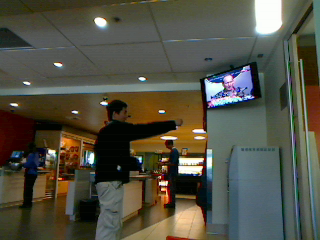} \\
   &  \emph{Boxing} &  \emph{Boxing}  &  \emph{Hand-clapping} & \emph{Hand-waving} &  \emph{Boxing} \\ \\
  \rotatebox{90}{\bf ~~~~~~~~UCF101} &
    \includegraphics[width=0.18\linewidth,height=2.4cm]{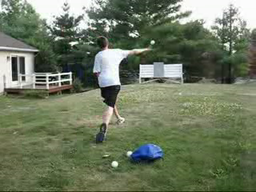} &
    \includegraphics[width=0.18\linewidth,height=2.4cm]{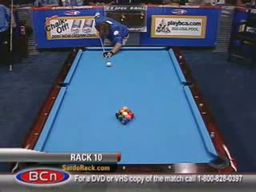} &
    \includegraphics[width=0.18\linewidth,height=2.4cm]{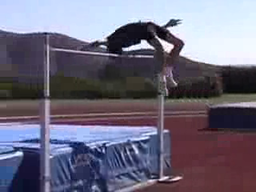} &
    \includegraphics[width=0.18\linewidth,height=2.4cm]{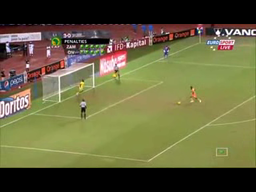} &
    \includegraphics[width=0.18\linewidth,height=2.4cm]{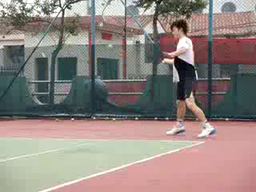}         \\
    & \emph{BaseballPitch} & \emph{Billiards} &  \emph{HighJump} &   \emph{Soccer penalty}  &  \emph{Tennis swing} \\  \\
    \end{tabular}
   \caption{Example video frames showing action classes from the UCF Sports, MSR-II and UCF101 datasets.}
\label{fig:db_exmp}}
\end{figure*}

\paragraph{Spatiotemporal refinement.} 
{A super-voxel and therefore a Tubelet capturing an actor/object can continue to extend further even after the action is completed as shown in the top row of Figure~\ref{tubelet_refinement}.
Tubelets are generated from super-voxels that generally follow an object or an actor and hence can be irregular in shape spatially, sometimes leading to sudden changes in the size of consecutive bounding boxes.
We propose to handle the above two problems of weak temporal localization and non-smooth spatial localization by temporal and spatial refinement.}

{\emph{Temporal refinement:}
In order to deal with the overly long Tubelets we propose to temporally sample or segment them. 
For this we devise a method that can segment each proposal into smaller sub-sequences with tighter temporal boundaries, without increasing the total number of proposals too much.
This temporal refinement is applied to one proposal at a time. Consider an action proposal of $B$ boxes (\ie, extending over $B$ frames) and $i^{th}$ box has $nrTraj(i)$ trajectories passing through it (where $i=1 \dots B$).
Now, we represent each box by two values, (a) relative location $=\frac{i}{B}$ and (b) relative motion content $=\frac{nrTraj(i)}{nrTraj_{max}}$. Here, $nrTraj_{max}$ is the maximum number of trajectories passing through any of the $B$ boxes. The boxes that have similar relative location and relative motion content are grouped together by clustering, such that the initial proposal is segmented into about fifteen sub-sequences. Then, very short proposals with temporal length less than thirty are filtered out. In practice, this increases the number of proposals by a factor ten. Therefore, we precede and follow temporal sampling by Overlap pruning, to restrict the total number of proposals. The impact of temporal refinement is shown in the second row of Figure~\ref{tubelet_refinement}}

%

{\emph{Spatial refinement:}
We apply spatial refinement of proposals, to steer the super-voxels closer to the shape of the action rather than the objects/actor and also to avoid sudden changes in sizes of bounding boxes and thus have smoother sequence of boxes. First, to align the boxes closer to action we modify them such that they are not void of motion trajectories at the boundaries. 
In each box, the minimum and maximum of $x$ and $y$ coordinates of intersecting trajectories are computed and the box is restricted to $[x_{min}-N, y_{min}-N, x_{max}+N, y_{max}+N]$.
Second, we apply weighted linear regression on width, height, $x$ and $y$ coordinates of the top left corner of the boxes. This is done over a local span of a few frames, typically a fifth of the proposal length. The impact of spatial refinement after temporal refinement is shown in the last row of Figure~\ref{tubelet_refinement}.}

\section{Datasets and Evaluation Criteria}
\label{tubelet_sec:setup}

\subsection{Datasets}
\paragraph{UCF Sports.} This dataset consists of 150 videos of actions extracted from sports broadcasts with realistic actions captured in dynamic and cluttered environments~\citep{Rodriguez:cvpr08}. This dataset is challenging due to many actions with large displacement and intra-class variability. Ten action categories are represented, for instance `diving', `swinging bench', `horse riding', etc.  We use the disjoint train-test split of videos (103 for training and 47 for testing)  suggested by \citep{tian_iccv11}.  The ground truth is provided as sequences of bounding boxes enclosing the actors. The area under the ROC curve (AUC) is the standard evaluation measure used, and we follow this convention.

\paragraph{MSR-II and KTH.} This dataset consists of 54 videos recorded in a crowded environment with many people moving in the background. Each video contains multiple actions of three types: `boxing', `hand clapping' and `hand waving'. An actor appears, performs one of these actions, and walks away. A single video has multiple actions (5-10) of different types, making the temporal localization challenging. Bounding subvolumes or cuboids are provided as the ground truth. Since the actors do not change their location, it is equivalent to a sequence of bounding boxes. The localization criterion is subvolume-based, so we follow \citep{Cao:cvpr10} and use the tight subvolume or cuboid enveloping Tubelet. Precision-recall curves and average precision (AP) are used for evaluation~\citep{Cao:cvpr10}. As standard practice, this dataset is used for cross-dataset experiments with KTH~\citep{kth_icpr04} as training set. 

\paragraph{UCF101.} The UCF101 dataset by \citep{ucf101} is a large action recognition dataset containing 101 action categories of which 24 are provided with localization annotations, corresponding to 3,204 videos. Each video contains one or more instances of same action class. It has large variations (camera motion, appearance, scale, etc.) and exhibits much diversity in terms of actions. Three train/test splits are provided with the dataset, we perform all evaluations on the first split with 2,290 videos for training and 914 videos for testing. Mean average precision is used for evaluation.

Example frames of some of the action classes are shown in Figure~\ref{fig:db_exmp} for each dataset.

\subsection{Evaluation criteria for action proposals} \label{tubelet_sec:evalLocalization}
To evaluate the quality of action proposals, we compute the upper bound on the localization accuracy, as previously done to evaluate the quality of object proposals~\citep{Jasper:selective}, by the Mean Average Best Overlap (MABO) and maximum possible recall. In this subsection, we extend these measures from objects in images to actions in videos. This requires measuring the overlap between two sequences of boxes instead of two boxes.

\paragraph{Overlap or localization score.} In a given video $V$ of $F$ frames comprising $m$ instances of different actions, the $i^{th}$ ground truth sequence of bounding boxes is given by $gt^i = (B_1^i, B_2^i, ... B_F^i)$. If there is no action of $i^{th}$ instance in frame $f$, then $B_f^i = \emptyset$. From the action proposals, the $j^{th}$ proposal formed by a sequence of bounding boxes is denoted as, $dt^j = (D_1^j, D_2^j, ... D_F^j)$. Let $OV_{i,j}(f)$ be the overlap between the two bounding boxes in frame, $f$, which is computed as intersection-over-union. The localization score between ground truth Tubelet $gt^i$ and a Tubelet $dt^j$ is given by: 
\begin{equation}
\label{eq:locscore}
S(gt^i,dt^j) = \frac{1}{|\Gamma|} \sum_{f \in \Gamma} OV_{i,j}(f), 
\end{equation} 
where $\Gamma$ is the set of frames where at least one of $B_f^i$, $D_f^j$ is not empty. This criterion generalizes the one proposed by \citep{tian_iccv11} by taking into account the temporal axis.

\paragraph{Mean Average Best Overlap (MABO).} The Average Best Overlap (ABO) for a given class $c$ is obtained by computing for each ground-truth annotation $gt^i \in G^c$, the best localization from the set of action proposals $T = \{dt^j | j=1 \dots m \}$:
\begin{equation}
  \textrm{ABO} = \frac{1}{|G^c|} \sum_{gt^i \in G^c} \max_{dt^j \in T} S(gt^i,dt^j).
\end{equation}
The mean ABO (MABO) summarizes the performance over all the classes. 

\paragraph{Maximum possible recall (\emph{Recall}).} Another measure for quality of proposals is maximum possible recall.
It is computed as the fraction of ground-truth actions with best overlap of greater than the overlap threshold ($\sigma$) averaged over action classes.
 We compute it with a very stringent localization threshold $\sigma=0.5$. 

Note that adding more proposals can only increase the MABO and Recall (scores are maintained if added proposals are not better). So, both MABO and Recall must be considered jointly with the number of proposals.

\paragraph{Action localization.}
An instance of action, $gt^i$, is considered to be correctly localized by an action proposal, $dt^j$, if the action is correctly predicted by the classifier and also the overlap/localization score is greater than the overlap threshold, \ie,  $S(gt^i,dt^j) > \sigma$.

\begin{table}[t]
\centering
{\small
 \tabcolsep=0.15cm
\begin{tabular}{lrrcr}
\toprule
{Segmenting} 								& {MABO} 	& {Recall} 		& {\# Super-voxels} & {Time (secs)}\\  \midrule
Video         								&  36.2     &  17.3       	& 862         & 379     \\
\imemotion maps                				&  48.6     &  53.2       	& 118         &  69     \\
\bottomrule
\end{tabular}}
\caption{Quality of initial super-voxels by applying the graph-based segmentation by \cite{xuCVPR12SuperVoxelEvaluation} on RGB video frames and on a sequence of \imemotion maps for the UCF Sports train set. We report MABO, Recall (at $\sigma=0.5$), number of initial super-voxels, and execution time in seconds. Note the competitive performance of super-voxel segmentation on \imemotion maps.}
\label{tab:segment}
\end{table}

\section{Experiments: Quality of Tubelets} \label{tubelet_sec:experiments}
In this section, we first analyze and evaluate the three stages of Tubelet extraction on the training set of the UCF Sports dataset. The initial step, super-voxel segmentation, is discussed in Subsection~\ref{tubelet_sec:segment}. Then, we evaluate different grouping functions over the initial set of super-voxels in Subsection~\ref{tubelet_sec:merging} and also show that segmenting iMotion maps is complementary to segmenting input video frames. 
In Subsection~\ref{tubelet_sec:pruning}, we evaluate the impact of spatiotemporal refinement and pruning on all three datasets. 
Finally, in Subsection~\ref{tubelet_sec:quality} we compare Tubelets with the state-of-the-art. 
We evaluate Tubelets with modern representations for action localization in Section~\ref{tubelet_sec:localize}.

\subsection{Super-voxel segmentation} \label{tubelet_sec:segment}
Here, we evaluate the graph-based segmentation of video and the graph-based segmentation of \imemotion maps. 
We set parameters as follows: $\sigma$ = 0.5, merging threshold of two nodes, $c=200$, minimum segment size $smin=500$, bigger $c$ and $smin$ would mean larger (and hence fewer) segments. 
In Table~\ref{tab:segment}, we compare the segmentation methods based on MABO, Recall, number of super-voxels and computation time. 
Segmentation of \imemotion maps leads to better results on all respects with higher MABO and Recall, fewer initial super-voxels and lower computation time. However, initial super-voxels from video segmentation are also important, as we will see in the next experiment. 

\begin{table}[t]
\centering
{\small
\begin{tabular}{lrrr} \toprule 
Super-voxel grouping 	        	& {MABO} 	& {Recall} 	&  {\#Proposals}      \\	\midrule	
\textbf{Single grouping function} 	& & &  			\\
\emph{Motion}        	 	 		& 56.2   	&    64.3   	& {\bf 299}         \\
\emph{Color}            			& 47.3   	&    42.0   	& 483                 \\
\emph{Texture}       				& 44.6   	&    36.2   	& 381                  \\
\emph{Size}            				& 47.8   	&    45.8   	& 918                  \\
\emph{Fill}                 		& 50.9   	&    50.4   	& 908      	           \\    
\emph{Motion+Size+Fill} 			& 57.2   	&    65.5   	& 719                  \\
\emph{Texture+Size+Fill}  		& 52.6   	&    57.5   	& 770                  \\
\emph{All-but-motion}      		& 53.4   	&    53.6   	& 672                  \\
\emph{All}               			& 58.1   	&    66.7   	& 656        	           \\ 	
 \midrule 
\textbf{Multiple grouping functions} & & &  			\\
{Union set, $\Phi$}				& {\bf 62.0} 	& {\bf 74.7}  	& 3,254              \\ 
	\bottomrule
\end{tabular}}
\caption{Evaluation of super-voxel groupings with \emph{video segmentation} on training set of UCF Sports.
Among the similarity measures, the ones based on \imemotion: \emph{Motion}, \emph{Motion+Size+Fill} and \emph{All} perform the best while generating a reasonable number of proposals. The union of the five selected grouping functions, $\Phi$, further increases the MABO and Recall.} 
\label{tubelet_table:mabo_gb}
\end{table}

\begin{table}[t]
\centering
{\small
\begin{tabular}{lrrr} \toprule 
	                
Super-voxel grouping 		  	& {MABO} 	& {Recall} 	&  {\#Proposals}      \\	\midrule	
\textbf{Single grouping function} 	& & &  			\\
\emph{Motion}          	 			& 52.9   	&  66.9    	& 90              \\
\emph{Color}            			& 51.1   	&  60.5    	& 93              \\
\emph{Texture}   					& 51.2   	&  62.5    	& {\bf 81}              \\
\emph{Size}              			& 52.2   	&  63.5    	& 158             \\
\emph{Fill}    					& 52.7   	&  61.9    	& 155             \\    
\emph{Motion+Size+Fill} 			& 54.2   	&  70.8    	& 129             \\
\emph{Texture+Size+Fill}  		& 53.9   	&  67.8    	& 145             \\
\emph{All-but-motion}      		& 54.5   	&  71.3    	& 127             \\
\emph{All}               			& 55.1   	&  74.5    	& 123             \\ 	
 \midrule 
\textbf{Multiple grouping functions} & & &  			\\
{Union set, $\Phi$}	 			& {\bf 56.8}   	& {\bf 77.0}    &       624             \\ 
	\bottomrule
\end{tabular}}
\caption{Evaluation of super-voxel groupings with \emph{segmentation of \imemotion maps} on the training set of UCF Sports. The grouping functions containing the \imemotion similarity measure again prove to be the most successful, though not as much as in Table~\ref{tubelet_table:mabo_gb}. The union set, $\Phi$, achieves a high MABO and Recall with only 624 proposals per video.} 
\label{tubelet_table:mabo_ime}
\end{table}

\subsection{Super-voxel grouping} \label{tubelet_sec:merging}
{We evaluate super-voxel groupings in Table~\ref{tubelet_table:mabo_gb} and Table~\ref{tubelet_table:mabo_ime} for video and \imemotion segmentations respectively. Nine grouping functions are considered that use one or more of the five similarity measures defined in Section~\ref{tubelet_sec:merge}: \emph{Motion}, \emph{Color}, \emph{Texture}, \emph{Size} and \emph{Fill}. Five of these use only one similarity measure, while the other four use multiple similarities. Here, \emph{All-but-motion} is Color+Texture+Size+Fill and \emph{All} is Motion+Color+Texture+Size+Fill, the rest are self-explanatory. We first evaluate these 9 grouping functions in both the tables.
In Table~\ref{tubelet_table:mabo_gb}, the best performing groupings are the ones that involve the \imemotion similarity measure: \emph{Motion}, \emph{Motion+Size+Fill} and \emph{All}. \emph{Motion} needs only 299 proposals per video to achieve a MABO of 56.2\% and Recall of 64.3\%. Note that it is much lower than the number of initial super-voxels (862) by the graph-based video segmentation. This is because \imemotion brings most of the motion content in fewer super-voxels and the majority of the resulting super-voxels are too small or have zero-motion, and hence are discarded.}

{After trying several combinations on the training set of UCF Sports, we select 5 best grouping functions: \emph{Motion}, \emph{Fill}, \emph{Motion+Size+Fill}, \emph{All-but-motion} and \emph{All}. Grouping the super-voxels from the five selected functions into a \emph{Union set}, $\Phi$} significantly increases the MABO and Recall to 62.0\% and 74.7\% respectively. Considering that a common localization score threshold ($\sigma$) used in the literature is 0.2 \citep{tian_iccv11,Yicong:sdpm}, these MABO values and Recall at $\sigma=0.5$ are very promising. 
{Thus obtained set of Tubelets with input video segmentation and \emph{Union set}, $\Phi$, is from now on referred to as \tubeletvid.}

Super-voxel groupings with \emph{segmentation of \imemotion maps} are evaluated in Table~\ref{tubelet_table:mabo_ime}.
Here, the grouping functions containing the \imemotion similarity measure again prove to be the most successful, though not as much as in the case of video segmentation. It is because by segmenting \imemotion maps motion information is already utilized to some extent. \emph{Fill} also leads to good MABO and Recall with just 155 proposals.
The union set, $\Phi$, achieves a good MABO of 56.8\% and Recall of 77.0\%, which even outperforms the Recall obtained with video segmentation by 2.3\%. Although the best MABO with segmentation of \imemotion maps is lower than that for video segmentation, the number of proposals required is only 624 on average, which is lower than the 3,254 proposals from video segmentation. This is a considerable reduction, which is in particular useful for long videos where the number of proposals can be high. Moreover segmenting \imemotion maps is faster, which is again of interest when operating on longer videos. 
{This set of Tubelets obtained by segmenting iMotion maps and \emph{Union set}, $\Phi$, is from here on referred to as \tubeletime.}

After analyzing segmentations from input video and \imemotion maps separately, we now combine the Tubelets from both, resulting proposal set denoted by \tubeletime $\cup$ \tubeletvid. As reported in Table~\ref{tubelet_table:mabo_comb}, the MABO increases up to 69.5\% and Recall reaches 93.6\%.
This is an improvement of $\sim$7\% in MABO and $\sim$16\% in Recall over the individual best of video and \imemotion segmentations.
{The experiments till this point are conducted on training set of UCF Sports. This validates the set of grouping functions, $\Phi$, and that the two Tubelet sets \tubeletime and \tubeletvid complement each other for localizing actions. We fix this setting for the experiments to follow.}

\begin{table}[t]
\centering
{\small
\begin{tabular}{lrrr}
\toprule
Super-voxel grouping 	  	& {MABO} 		& {Recall} 		& {\#Proposals}\\	\midrule
\emph{Motion}          			& 63.9   		&    80.9   	& {\bf 390}			\\
\emph{Fill}            			& 62.2   		&    77.5   	&  1,062		\\
\emph{Motion+Size+Fill}   	& 65.1   		&    86.4   	&   848			\\
\emph{All-but-motion}			& 65.0			&    86.0		&   799			\\
\emph{All}						& 66.6			&    91.3		&   779			\\	\midrule
{Union set, $\Phi$}			& {\bf 69.5}	&  {\bf 93.6}	&  3,878 		\\	\bottomrule
\end{tabular}}
\caption{Combining of Tubelets from video segmentation and \imemotion segmentation, \tubeletvid $\cup$ \tubeletime. Numbers are reported for the five selected grouping functions as well as their union set, $\Phi$. The combination leads to significant improvement of MABO and Recall, showing the two sets of Tubelets from two video segmentations complement each other.}
\label{tubelet_table:mabo_comb}
\end{table}

\subsection{Pruning and spatiotemporal refinement} \label{tubelet_sec:pruning}
In this section, we evaluate the impact of pruning and spatiotemporal refinement on the quality of action proposals of UCF Sports, MSR-II and UCF101. The validation for grouping functions and segmentation is already done on the training set of UCF Sports. Now, we report results when considering \emph{all} the videos of these three datasets, to be comparable with the numbers reported by other methods. Before moving to results, we provide the implementation details of pruning and spatiotemporal refinement.

\paragraph{Implementation details.}
For motion pruning we set $P=50$, so that at least fifty proposals are retained from each video. Also, motion pruning is only applied to \tubeletvid, since proposals from \tubeletime are expected to have enough motion content. 
Overlap pruning is similar to non-maximum suppression, but applied without classification scores and therefore can affect the recall. To minimize its impact on Recall, we set a high overlap threshold of $0.8$ for overlap based pruning. 
For spatial refinement, we set $N$ equal to $5\%$ of the frame width.

\begin{table}[t]
\centering
{\small
\begin{tabular}{lrrr}
\toprule
 			  	   & {MABO}     & { Recall}	& {\#Proposals}         \\ 
\midrule
\tubeletvid $\cup$ \tubeletime     & 69.3  & 93.5    & 3,432 	\\
+ Motion pruning 	           & 69.3  & 93.5    & 884   	\\
+ Overlap pruning		   & 67.5  & 90.5    & 289        \\ 
+ Spatial refinement		   & 67.5  & 91.9    & 289     \\
  \bottomrule
\end{tabular}}
\caption{Impact of pruning and spatial refinement of Tubelets on UCF Sports: Even after motion pruning the MABO and Recall are maintained with only $\sim26$\% of proposals. With overlap pruning the number of proposals goes down further to $\sim8$\% of the original number, with a small loss in MABO and Recall scores. The loss is compensated by spatial refinement of Tubelets.}
\label{tubelet_table:prune_refine_ucfs}
\end{table}

\paragraph{UCF Sports.}
{In Table~\ref{tubelet_table:prune_refine_ucfs}, we evaluate the impact of pruning and spatial refinement on MABO, Recall and the average number of proposals per video for UCF Sports dataset. The results for \tubeletvid $\cup$ \tubeletime for all 150 videos of UCF Sports is similar to that on its train set. Now, with motion pruning there is no loss of  MABO and Recall while only $\sim26$\% of original proposals are used. Further, with overlap pruning number of proposals further goes down to $\sim8$\% of original number with a small loss in MABO and Recall. Finally, with spatial refinement of Tubelets there is small improvement of Recall. Altogether, with pruning and spatial refinement we are able to decrease the number action proposals by a factor 12 with only a modest loss in MABO and Recall.}

\begin{table}[h]
\centering
{\small
\begin{tabular}{lrrrr} \toprule
Localization              			& MABO 		& Recall  	& \#Proposals    \\ \midrule
Spatiotemporal					& 28.2		& 2.2		& 2,342         \\          
Spatial only          				& 60.9 		& 81.3      & 2,342     	\\ 
\bottomrule
\end{tabular}}
\caption{Spatial localization versus spatiotemporal localization on untrimmed videos of MSR-II: Spatial only localization leads to much better Recall, which indicates that the low Recall is due to weak temporal localization. This calls for temporal refinement of Tubelets.}
\label{tubelet_table:spatiotemp_loc}
\end{table}

\begin{table}[t]
\centering
{\small
\begin{tabular}{lrrr}
\toprule
	 	  			   & {MABO}     & { Recall}	& {\#Proposals}         \\ 
\midrule
\tubeletvid $\cup$ \tubeletime             & 36.9  &  5.1    & 25,962 	\\
+ Motion pruning		           & 36.7  &  5.1    &  6,560	\\
+ Temporal refinement			   & 46.0  & 35.2    & 7,287	\\
+ Spatial refinement			   & 48.9  & 47.4    & 7,287    \\
  \bottomrule
\end{tabular}}
\caption{Impact of pruning and spatial refinement of Tubelets on MSR-II: Pruning by motion maintains the MABO and Recall while reducing the proposals to only a quarter of the initial set. Temporal refinement has a positive impact on proposal quality with Recall increased by 30\%. Finally, with spatial refinement another improvement of $\sim12$\% is achieved. Spatiotemporal refinement is important for this dataset.}
\label{tubelet_table:prune_refine_msr}
\end{table}

\paragraph{MSR-II.}
The MSR-II dataset has untrimmed videos with multiple instances of different types of actions in the same video. This poses additional challenges for temporal localization, which is experimentally illustrated in Table~\ref{tubelet_table:spatiotemp_loc}. 
The table reports MABO and Recall for Tubelet set \tubeletvid after motion pruning for spatiotemporal localization and also spatial-only localization. Overlap score for spatiotemporal case is computed according to Equation~\ref{eq:locscore} as done in all other results. For spatial localization, we compute only for the frames where ground-truth proposal is present, \ie, we do not penalize overlap score for temporal misalignment. MABO doubles and the Recall shoots from 2.2\% to 81.3\% for spatial-only localization, which means that our Tubelets very well locate the actions spatially but extends to the frames where there in no action of interest. This is due the tendency of super-voxels to continue to cover the actor even when the action is completed. We overcome this limitation by temporal refinement.

{In Table~\ref{tubelet_table:prune_refine_msr}, in addition to pruning and spatial refinement, we also report for temporal refinement to improve temporal localization. First, motion pruning maintains the MABO and Recall while reducing the number of proposals to only a quarter of initial number. This pruning needs to precede temporal refinement to limit the number of proposals.
Second, temporal refinement leads to a massive improvement of 30.1\% in Recall and 9.3\% in MABO. Note that temporal refinement also includes overlap pruning to filter-out newly added very similar proposals. Also, to limit the number of proposals temporal refinement is exclusively applied to `\tubeletvid + Motion pruning', which means only overlap pruning is applied to `\tubeletime + Motion pruning'. 
Finally, with spatial refinement another huge improvement of $\sim12$\% is achieved in Recall along with $\sim3$\% improvement in MABO. 

Overall, we achieve an improvement of 12\% of MABO and 42.3\% of Recall while decreasing the number of proposals by about 72\% compared to the initial set, \tubeletvid $\cup$ \tubeletime. 
The gain due to temporal refinement is easy to understand for this dataset of untrimmed videos. However, we also get impressive boost by spatial refinement that is much more than we get for the other two datasets. We attribute this to the exploitation of information from motion trajectories, which is paramount for MSR-II as noted before in~\cite{Gemert_BMVC15,chen2015action}.
}

\paragraph{UCF101.}
{In Table~\ref{tubelet_table:prune_refine_ucf101}, we report the impact of pruning and spatial refinement on MABO, Recall and the average number of proposals per video for UCF101 dataset.
Motion pruning also works well on the 3,204 videos of UCF101, compressing the number of proposals by a factor of four, while maintaining MABO and Recall. Further, with overlap pruning number of proposals further goes down to $\sim9$\% of original number with a small loss in MABO and Recall. With favourable spatial refinement, eventually, final set of Tubelets achieve same performance as by \tubeletvid $\cup$ \tubeletime, but with about 10 times fewer proposals.}

\begin{table}[t]
\centering
{\small
\begin{tabular}{lrrr}
\toprule
	 	  		& {MABO}     & { Recall}	& {\#Proposals}         \\ 
\midrule
\tubeletvid $\cup$ \tubeletime             & 42.6  & 33.4    & 5,410 	\\
+ Motion pruning		           & 41.7  & 32.5    & 1,298  	\\
+ Overlap pruning			   & 40.9  & 30.6    & 472        \\ 
+ Spatial refinement			   & 42.3  & 32.8    & 472     \\
  \bottomrule
\end{tabular}}
\caption{{Impact of pruning and spatial refinement of Tubelets on UCF101:} Motion pruning leads to $\sim1$\% loss in MABO and Recall while filtering out 75\% of the proposals. With overlap pruning the number of proposals goes down further to $\sim9$\% of the original number with a small loss in MABO and Recall. This loss is compensated by spatial refinement leading to the same performance with ten times fewer proposals.}
\label{tubelet_table:prune_refine_ucf101}
\end{table}

\subsection{Comparison with state-of-the-art methods} \label{tubelet_sec:quality}
In Table~\ref{tab:methodsComp}, we compare our Tubelets with alternative unsupervised action proposals from the literature. With a relatively small set 289 proposals we outperform all the other approaches on UCF Sports. 
{On MSR-II, we outperform the previous best approach of \cite{Gemert_BMVC15}. It is interesting to note the improvement in MABO and Recall over the initial version of our approach \citep{Jain:tubelets}, indicating the value of spatiotemporal refinement and pruning. On UCF101, we achieve MABO and Recall comparable to the method of~\cite{Gemert_BMVC15}, be it that we need five times less proposals. Overall, Tubelets provides state-of-the-art quality while balancing the number of proposals. Next we evaluate the action localization abilities of Tubelets when combined with modern representations. }

\begin{table}[t]
\centering
    \scalebox{0.95}{
\begin{tabular}{lrrr} \toprule
  & MABO & Recall  & \#Proposals \\ \midrule
\textbf{UCF Sports}  & & & \\ 
\cite{Jain:tubelets}                                                       & 62.7 & 78.7 & 1,642 \\
\cite{oneataECCV14spatemprops}                            & 55.6 & 68.1 & 3,000 \\
\cite{Gemert_BMVC15}                                              & {64.2} & {89.4} & 1,449 \\
Tubelets                                                             & \textbf{67.5} & \textbf{91.9} & \textbf{289} \\ \midrule
\textbf{MSR-II}  & & &  \\ 
\cite{Jain:tubelets}                                                     & 34.8 & 3.0 & 4,218 \\
\cite{Gemert_BMVC15}                                  		& {47.9} & {44.3} & \textbf{6,706} \\
Tubelets                                                                    & \textbf{48.9}        & \textbf{47.4}         & 7,287 \\ \bottomrule
\textbf{UCF 101}  & & & \\ 
\cite{Gemert_BMVC15}                                              & {40.0} & \textbf{35.5} & 2,299 \\
Tubelets                                                                               & \textbf{42.3} & {32.8} & \textbf{472} \\ \midrule
\end{tabular}
}
\caption{Comparing quality of action proposals against state-of-the-art. Our Tubelets outperform all other approaches on these three datasets with a modest number of proposals. Our Recall on UCF101 is slightly behind the approach of \cite{Gemert_BMVC15}, be it they use five times more proposals.} 
\label{tab:methodsComp}
\end{table}

\section{Experiments: Action localization}
\label{tubelet_sec:localize}
In this section we evaluate our approach for action localization UCF Sports, MSR-II and UCF101.
 For positive training examples, we use the ground-truth and our Tubelets that have localization score greater than $0.7$ with the ground-truth. Negative samples are randomly selected by considering Tubelets whose overlap with ground-truth is less than $0.15$. This scheme is followed for UCF Sports and UCF101. In case of MSR-II cross-dataset evaluation is employed, the training samples consist of the clips from KTH dataset while testing is performed on the Tubelets from the videos of  MSR-II.
We apply power normalization followed by $\ell_2$ normalization before training with a linear SVM. One round of retraining on ``hard-negatives" was enough as additional rounds did not improve performance further. Again there is no retraining in case of MSR-II, only initial classifier trained on videos from KTH dataset are used.

We first give details of the representations used to encode each Tubelet and show their impact on the UCF Sports dataset. Then, we compare our action localization results with the state-of-the-art methods on each of the three datasets.

\subsection{Tubelet representations} \label{sec:encoding}
We capture motion information by the four local descriptors computed along the improved trajectories~\citep{wang:imptraj13}. To represent the local descriptors, we use bag-of-words or Fisher vectors. A Tubelet is assigned the trajectories that have more than half of there points inside the Tubelet.
For the third representation we use features from a Convolutional Neural Network layer and average pool them over the frames. Below we explain these three representations.

\begin{figure}[t]
\centering
  \includegraphics[width=1\linewidth]{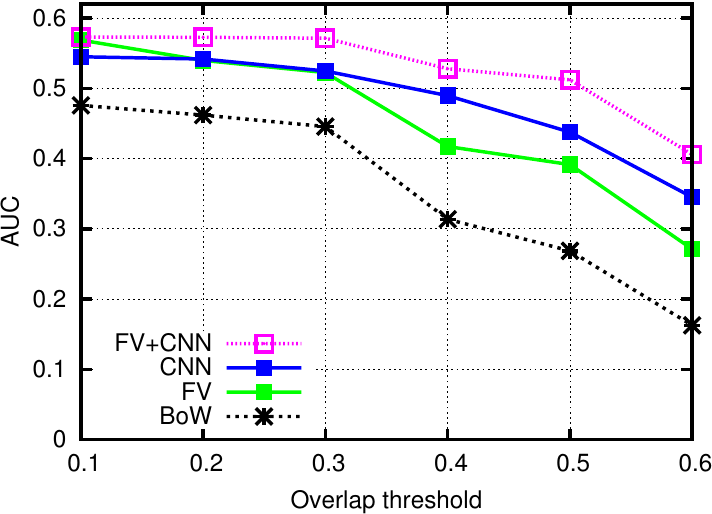}
  \caption{Comparing representations: Bag-of-words, Fisher vector and CNN features on UCF Sports, performance is measured by AUC for $\sigma$ from 0.1 to 0.6, following~\citep{Yicong:sdpm}. The best AUC is obtained when both Fisher vector and CNN features are combined for the Tubelet representation.
  }
  \label{tubelet_fig:ucfs_rep}
\end{figure}

\paragraph{Bag of words (BoW)}.
The local descriptors are vector quantized and pooled into a bag-of-words histogram. We set the vocabulary size to $K=500$. This is the least expensive (and expressive) of the three representations.

\paragraph{Fisher vectors (FV)}.
We first apply PCA on the local descriptors and reduce the dimensionality by a factor of two. Then 256,000 descriptors are selected at random from a training set to estimate a Gaussian Mixture Model with $K$ (= 128) Gaussians. Each video is then represented by $2DK$ dimensional Fisher vector, where $D$ is the dimension of the descriptor after PCA. Finally, we apply power and $\ell_2$ normalization to the Fisher vector as suggested in~\citep{PSM10}. The feature computation is reasonably efficient but the memory requirement would be a bottleneck if the number of proposals are high (\eg $>5000$). Fisher vectors have been used for temporal action localization by~\citep{oneata:cvpr_14} and for spatiotemporal action localization by \cite{Gemert_BMVC15}.

\paragraph{Convolutional neural network (CNN)}.
We use an in-house implementation of GoogLeNet~\citep{googlenet}, trained on ImageNet over 15k object categories~\citep{Jain:15kobjects} without fine-tuning. 
The features are extracted from the fully-connected layer (before softmax2) of the network, which is a $1024$ dimensional vector to represent a bounding box in a frame. Since a Tubelet is a sequence of bounding boxes, the final representation for it is obtained by averaging the feature vectors for the sampled frames (2 frames per second). Here, the memory requirement is limited, and feature computation is the costly operation, motivating the need for a compact set of action proposals.

\paragraph{Comparing representations}.
We now analyze the impact of the above three Tubelet representations on the UCF Sports dataset, following the process described in Section~\ref{tubelet_sec:evalLocalization}. Following popular practice, we use area under ROC curve (AUC) as the evaluation measure, as common for this dataset.
Figure~\ref{tubelet_fig:ucfs_rep} compares the performance of the various Tubelet representations for a varying overlap threshold. We observe a clear improvement when moving from BoW to FV, to CNN and eventually the combination of FV and CNN, especially for higher thresholds ($\sigma \geq 0.4$).

\begin{figure}[t]
\centering
  \includegraphics[width=1\linewidth]{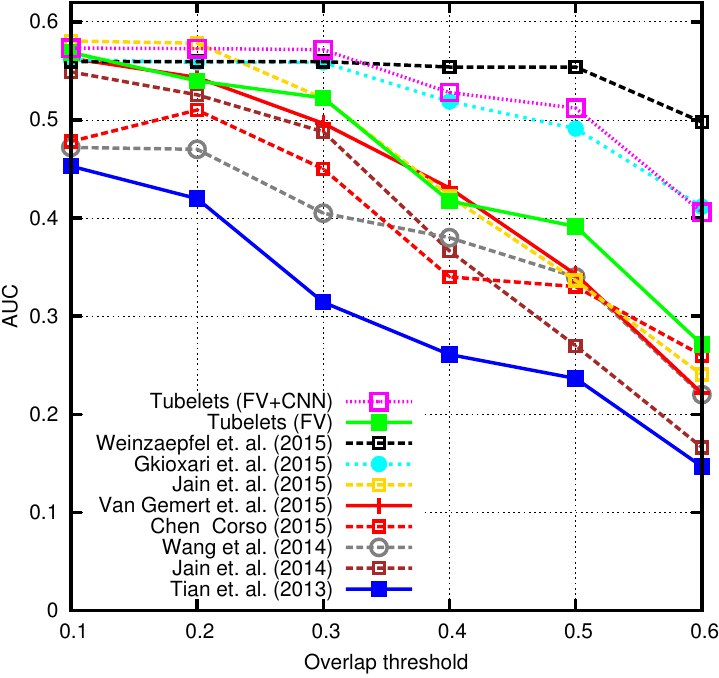}
  \caption{Comparison with state-of-the-art methods on UCF Sports, performance is measured by AUC for $\sigma$ from 0.1 to 0.6.}
  \label{tubelet_fig:ucfs_soa}
\end{figure}

\subsection{Comparison with state-of-the-art methods} \label{locexp:soa}
We now compare our approach with state-of-the-art methods on the three datasets. 

\begin{figure*}[t]
\begin{minipage}{1\linewidth}
\begin{minipage}{0.24\linewidth}
\includegraphics[width=0.98\linewidth]{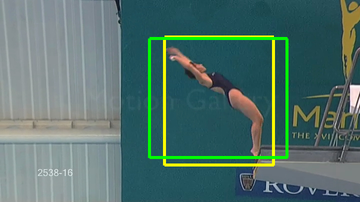} \smallskip \\
\includegraphics[width=0.98\linewidth]{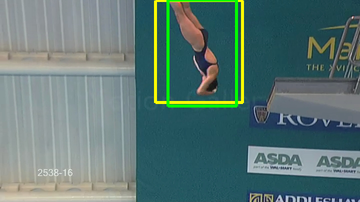}       \smallskip \\
\includegraphics[width=0.98\linewidth]{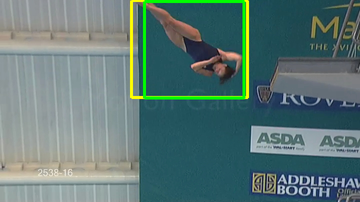}   
\centering (a) Diving
\end{minipage}
\hfill
\begin{minipage}{0.24\linewidth}
\includegraphics[width=0.98\linewidth]{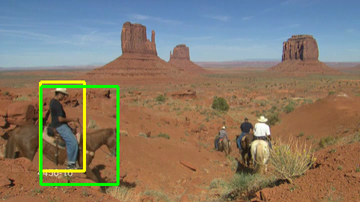} \smallskip \\
\includegraphics[width=0.98\linewidth]{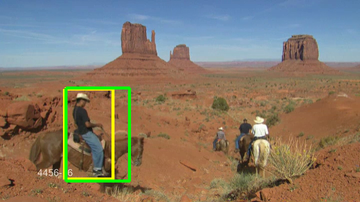} \smallskip  \\
\includegraphics[width=0.98\linewidth]{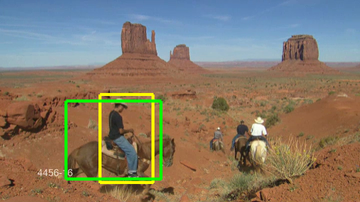}
\centering (b) Riding-Horse
\end{minipage}
\hfill
\begin{minipage}{0.24\linewidth}
\includegraphics[width=0.98\linewidth]{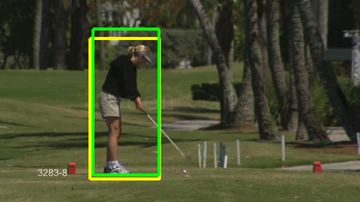} \smallskip \\
\includegraphics[width=0.98\linewidth]{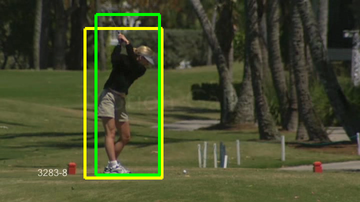} \smallskip    \\
\includegraphics[width=0.98\linewidth]{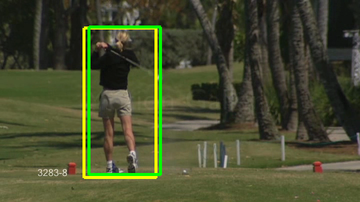}
\centering (c) Golf
\end{minipage}
\hfill
\begin{minipage}{0.24\linewidth}
\begin{center}
\includegraphics[width=0.98\linewidth]{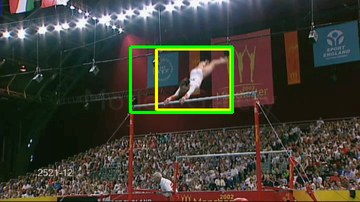}         \smallskip \\
\includegraphics[width=0.98\linewidth]{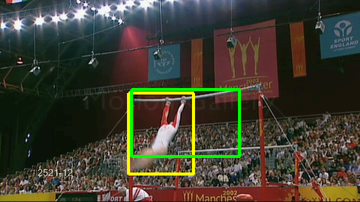}        \smallskip      \\
\includegraphics[width=0.98\linewidth]{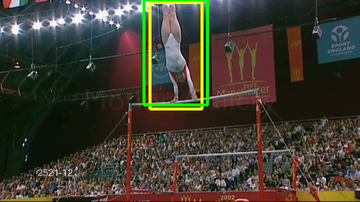}
\centering (d) Swinging-Bar
\end{center}
\end{minipage} ~ \smallskip \\
\end{minipage}          \\

\begin{minipage}{1\linewidth}
\begin{minipage}{0.24\linewidth}
\includegraphics[height=2.6cm,width=0.98\linewidth]{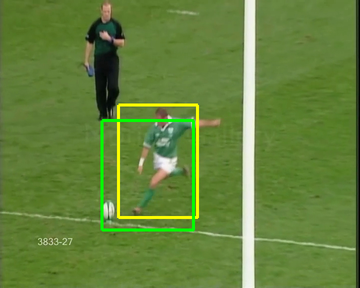} \smallskip \\
\includegraphics[height=2.6cm,width=0.98\linewidth]{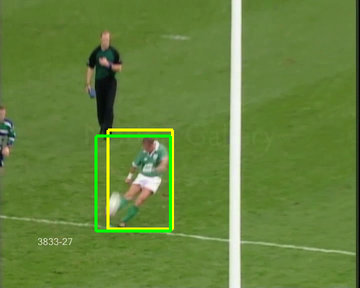}       \smallskip \\
\includegraphics[height=2.6cm,width=0.98\linewidth]{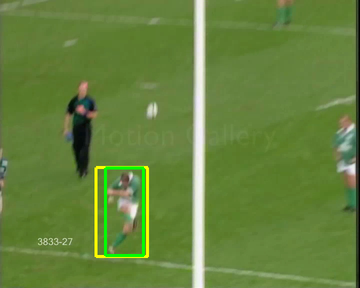}
\centering (e) Kicking
\end{minipage}
\hfill
\begin{minipage}{0.24\linewidth}
\includegraphics[height=2.6cm,width=0.98\linewidth]{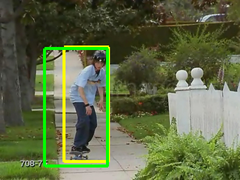} \smallskip \\
\includegraphics[height=2.6cm,width=0.98\linewidth]{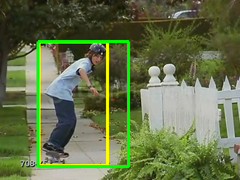} \smallskip  \\
\includegraphics[height=2.6cm,width=0.98\linewidth]{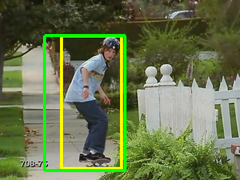}
\centering (f) Skating
\end{minipage}
\hfill
\begin{minipage}{0.24\linewidth}
\includegraphics[height=2.6cm,width=0.98\linewidth]{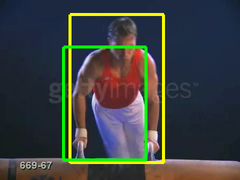} \smallskip \\
\includegraphics[height=2.6cm,width=0.98\linewidth]{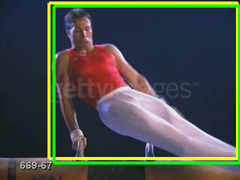} \smallskip    \\
\includegraphics[height=2.6cm,width=0.98\linewidth]{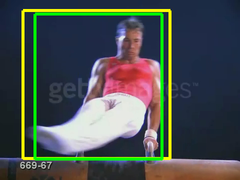}
\centering (g) Swinging-Bench
\end{minipage}
\hfill
\begin{minipage}{0.24\linewidth}
\begin{center}
\includegraphics[height=2.6cm,width=0.98\linewidth]{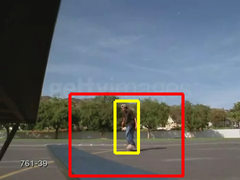}         \smallskip \\
\includegraphics[height=2.6cm,width=0.98\linewidth]{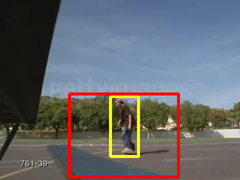}        \smallskip      \\
\includegraphics[height=2.6cm,width=0.98\linewidth]{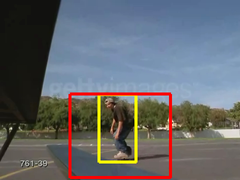}
\centering (h) Skating
\end{center}
\end{minipage} ~ \smallskip \\
\end{minipage}

\caption{Localization results shown as a sequence of bounding boxes (UCF-Sports): Ground-truth is shown in yellow, correctly localized detections in green and poorly localized ones in red. Caption below each sequence reports the class detected.}
\label{tubelet_fig:dets_ucf}
\end{figure*}

\paragraph{UCF Sports}.
In Figure~\ref{tubelet_fig:ucfs_soa}, we compare the performance of our method with the best reported results in the literature. In \citep{Jain:15kobjects}, the previous version of Tubelets were represented with FV and CNN features, hence for comparison we use Tubelets represented with FV+CNN.
The boost over \cite{Jain:15kobjects}, relying on segmentation of video frames only, shows the importance of segmenting \imemotion maps as well.
Tubelets represented with FV+CNN is competitive to the methods of \cite{gkioxariCVPR15actionTubes} and \cite{Weinzaepfel_iccv15} and outperforms all other approaches. Since \cite{Gemert_BMVC15} uses only the FV representation, for fair comparison we also include Tubelets with a FV representation, which does better for most of the thresholds. Figure~\ref{tubelet_fig:dets_ucf} shows some examples of action localizations from UCF Sports.

\begin{figure*}[t]
\begin{minipage}{1\linewidth}
\begin{minipage}{0.24\linewidth}
\includegraphics[width=0.98\linewidth]{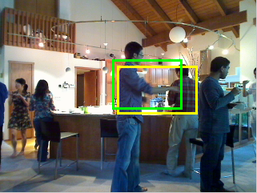} 	\smallskip \\
\includegraphics[width=0.98\linewidth]{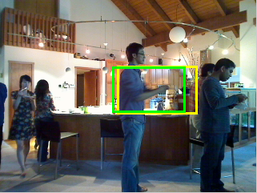}     \smallskip \\
\includegraphics[width=0.98\linewidth]{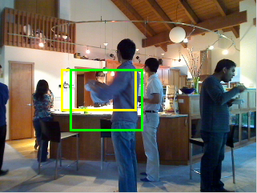}     \smallskip \\
\includegraphics[width=0.98\linewidth]{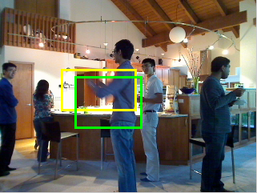}
\centering (a) Boxing
\end{minipage}
\hfill
\begin{minipage}{0.24\linewidth}
\includegraphics[width=0.98\linewidth]{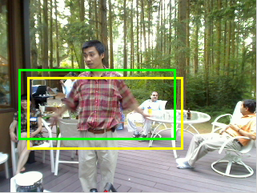} \smallskip \\
\includegraphics[width=0.98\linewidth]{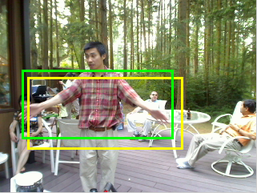} \smallskip  \\
\includegraphics[width=0.98\linewidth]{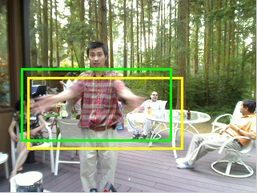} \smallskip  \\
\includegraphics[width=0.98\linewidth]{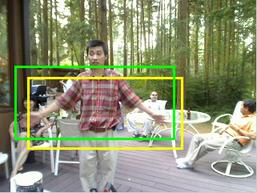} \smallskip  \\
\centering (b) HandClapping
\end{minipage}
\hfill
\begin{minipage}{0.24\linewidth}
\includegraphics[width=0.98\linewidth]{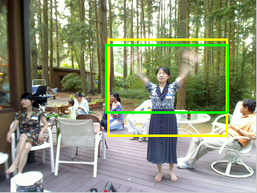} \smallskip    \\
\includegraphics[width=0.98\linewidth]{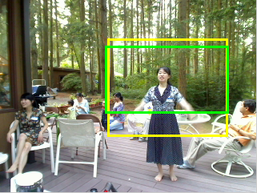} \smallskip    \\
\includegraphics[width=0.98\linewidth]{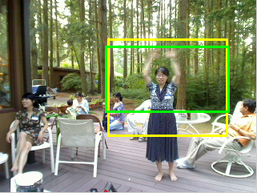} \smallskip    \\
\includegraphics[width=0.98\linewidth]{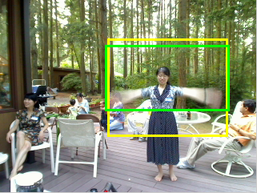} \smallskip    \\
\centering (c) HandWaving
\end{minipage}
\hfill
\begin{minipage}{0.24\linewidth}
\begin{center}
\includegraphics[width=0.98\linewidth]{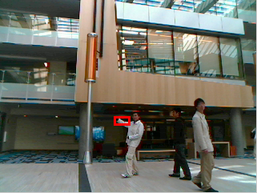}         \smallskip \\
\includegraphics[width=0.98\linewidth]{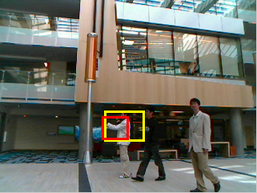}        \smallskip      \\
\includegraphics[width=0.98\linewidth]{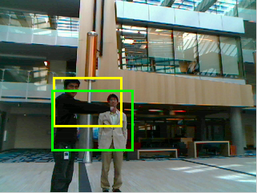}        \smallskip      \\
\includegraphics[width=0.98\linewidth]{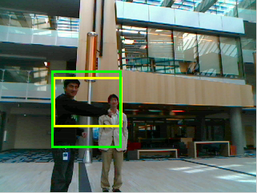}
\centering (d) Boxing
\end{center}
\end{minipage} ~ \smallskip \\
\end{minipage}          \\

\caption{Localization results shown as a sequence of bounding boxes (MSR-II): Ground-truth is shown in yellow, correctly localized detections in green and poorly localized ones in red. Two instances of `Boxing' being correctly localized are shown in the first column. The middle two columns show successful results for `Clapping' and `Waving' actions. Last column shows a failure case of poor localization of an instance of `Boxing', while the second instance in the video is localized well.}
\label{tubelet_fig:dets_msr}
\end{figure*}

\paragraph{MSR-II}.
This dataset is designed for cross-dataset evaluation. Following standard practice, we train on KTH dataset and test on MSR-II. While training for one class, the videos from other classes are used as the negative set.
We use the FV representation to be more comparable with the competitive work of~\citep{Gemert_BMVC15}, which also generates action proposals in an unsupervised manner like Tubelets.
In Table~\ref{tubelet_table:msr_ap}, we compare with several state-of-the-art methods; mean average precision (mAP) along with the APs for the three classes are reported. Following the usual practice on this dataset we report results for an overlap threshold of $0.125$.
{Apart from \cite{chen2015action}, our approach outperforms all other methods by $5$\% of mAP or more. 
\cite{chen2015action} very well utilizes information from motion trajectories and samples action proposals by clustering over a space-time trajectory graph. 
Motion trajectory based approaches are particularly well-suited for MSR-II dataset, as observed with our spatiotemporal refinement of Tubelets and also in~\citep{Gemert_BMVC15}.
Similarly, the approach of \cite{chen2015action} that is mainly focused on trajectories lead to excellent performance on MSR-II but its performance on UCF Sports is modest (Figure~\ref{tubelet_fig:ucfs_soa}).}
{Finally, compared to the Tubelets in~\cite{Jain:tubelets}, we improve mAP by 24.5\%. Again, we claim the importance of using both input video frames and \imemotion maps for segmentation and spatiotemporal refinement of Tubelets. Figure~\ref{tubelet_fig:dets_msr} shows some examples of localizations for MSR-II.}

\rowcolors{2}{white}{gray!10}

\begin{table}[t]
\centering
{
 \tabcolsep=0.15cm
\begin{tabular}{lccc|c} \toprule
\textbf{Method}   				& \textbf{Boxing}	& \textbf{Clapping}	& \textbf{Waving}	& \textbf{mAP}	\\ 	\midrule
~\cite{Cao:cvpr10}			&       17.5	   		&       13.2  		&    26.7		&	19.1		\\	
~\cite{Yicong:sdpm}			&       38.9	   		&       23.9		&    44.7   	     	 &	35.8	\\	
~\cite{Jain:tubelets} 	        	&       46.0			&       31.4  		&    {85.8}  	 & 54.4      	\\ 	
~\cite{Yuan:pami11}			&	 64.9   		&       43.1    	&     64.9 		 & 55.3 \\	
~\cite{Wang_dynamicPoslets}		&       41.7   			&      	50.2     	&     80.9 		 & 57.6 \\
~\cite{yuCVPR15fasttubes}		&       {67.4} 			&   	46.6 		&     69.9 		 & 61.3 \\
~\cite{mosabbebACCV2014weakly}		&       72.4   			&      	56.9     	&     81.1 		 & 70.1 \\
~\cite{Gemert_BMVC15}			&       67.0			&	{ 78.4}	&     74.1		&  {73.2}	\\
~\cite{chen2015action}			&       {\bf 94.4}		&	73.0		&     {\bf 87.7}	&  {\bf 85.0}	\\
\midrule
Tubelets 				&	72.4			&	{\bf 79.9}		&    84.4	&	78.9	\\	
\bottomrule
\end{tabular}}
\caption{Comparison with state-of-the-art methods on MSR-II: Average precision (AP) and mean AP are reported.}
\label{tubelet_table:msr_ap}
\end{table}

\begin{figure}[t]
\centering
 \includegraphics[width=0.95\linewidth]{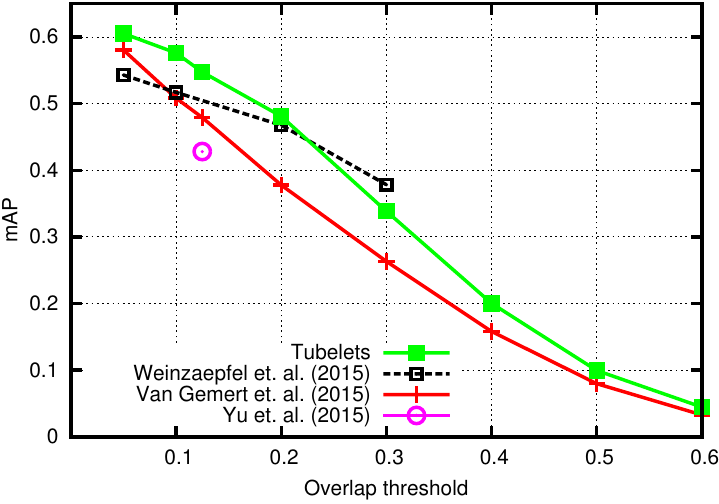}
  \caption{Comparison with state-of-the-art methods on UCF101: Tubelets are obtained using the selected five grouping functions and represented with FV. Performance is measured by mAP for $\sigma$ from 0.1 to 0.6.}
  \label{tubelet_fig:ucf101_comp}
\end{figure}

\paragraph{UCF101}.
UCF101 is much larger than the other two datasets, with 24 action classes, and is currently the most challenging dataset for classification of proposals. 
Again, we represent Tubelets with FV following~\citep{Gemert_BMVC15}. In Figure~\ref{tubelet_fig:ucf101_comp}, we report mAPs for different overlap thresholds and compare Tubelets with three other approaches that report results on UCF101 dataset. 
Despite the use of human detection, the approach by \cite{yuCVPR15fasttubes} is about $10$\% behind our method for an overlap threshold of $0.125$. 
\cite{Weinzaepfel_iccv15} uses bounding-box level action class supervision while generating proposals. Despite their additional supervision and use of two-stream CNN features, we achieve better mAP for 3 out of 4 overlap thresholds. The only other approach that uses proposals generated in an unsupervised manner, as we do, is APT by~\citep{Gemert_BMVC15}. Tubelets outperform their approach while requiring only about a fifth of proposals (see Table~\ref{tab:methodsComp}).

\begin{figure*}[t]
\begin{minipage}{1\linewidth}
\begin{minipage}{0.24\linewidth}
\includegraphics[height=2.6cm,width=0.98\linewidth]{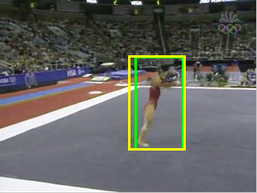}         \smallskip \\
\includegraphics[height=2.6cm,width=0.98\linewidth]{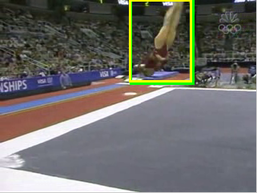}        \smallskip      \\
\includegraphics[height=2.6cm,width=0.98\linewidth]{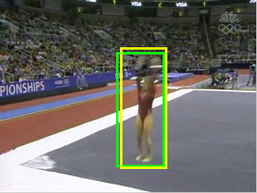}
\centering (a) Floor Gymnastics
\end{minipage}
\hfill
\begin{minipage}{0.24\linewidth}
\includegraphics[height=2.6cm,width=0.98\linewidth]{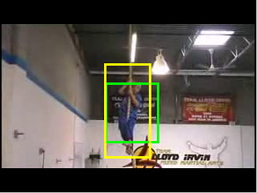}         \smallskip \\
\includegraphics[height=2.6cm,width=0.98\linewidth]{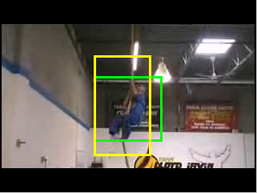}        \smallskip      \\
\includegraphics[height=2.6cm,width=0.98\linewidth]{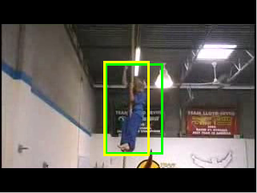}
\centering (b) Rope Climbing
\end{minipage}
\hfill
\begin{minipage}{0.24\linewidth}
\includegraphics[height=2.6cm,width=0.98\linewidth]{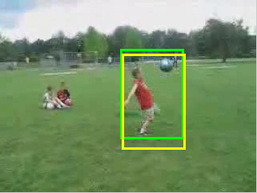}         \smallskip \\
\includegraphics[height=2.6cm,width=0.98\linewidth]{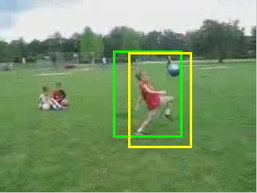}        \smallskip      \\
\includegraphics[height=2.6cm,width=0.98\linewidth]{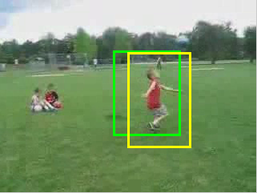}
\centering (c) Soccer Juggling
\end{minipage}
\hfill
\begin{minipage}{0.24\linewidth}
\begin{center}
\includegraphics[height=2.6cm,width=0.98\linewidth]{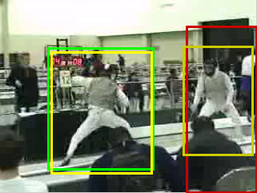}         \smallskip \\
\includegraphics[height=2.6cm,width=0.98\linewidth]{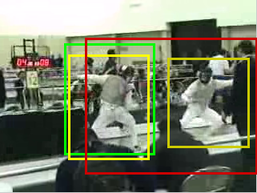}        \smallskip      \\
\includegraphics[height=2.6cm,width=0.98\linewidth]{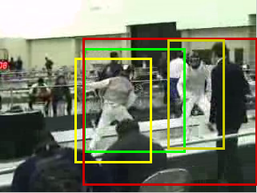}
\centering (d) Fencing
\end{center}
\end{minipage} ~ \smallskip \\
\end{minipage}          \\

\begin{minipage}{1\linewidth}
\begin{minipage}{0.24\linewidth}
\includegraphics[height=2.6cm,width=0.98\linewidth]{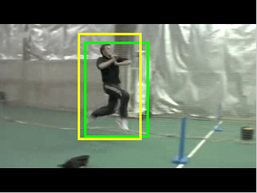}         \smallskip \\
\includegraphics[height=2.6cm,width=0.98\linewidth]{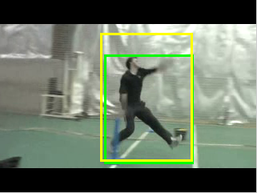}        \smallskip      \\
\includegraphics[height=2.6cm,width=0.98\linewidth]{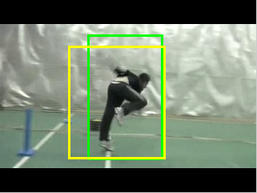}
\centering (e) Cricket Bowling
\end{minipage}
\hfill
\begin{minipage}{0.24\linewidth}
\includegraphics[height=2.6cm,width=0.98\linewidth]{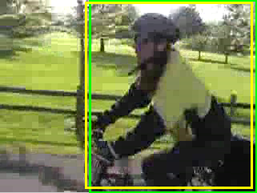}         \smallskip \\
\includegraphics[height=2.6cm,width=0.98\linewidth]{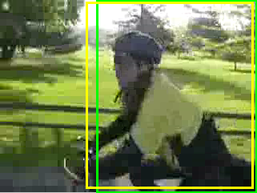}        \smallskip      \\
\includegraphics[height=2.6cm,width=0.98\linewidth]{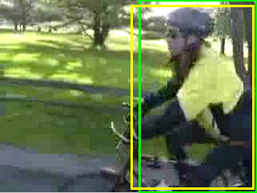}
\centering (f) Biking
\end{minipage}
\hfill
\begin{minipage}{0.24\linewidth}
\includegraphics[height=2.6cm,width=0.98\linewidth]{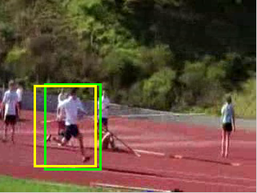}         \smallskip \\
\includegraphics[height=2.6cm,width=0.98\linewidth]{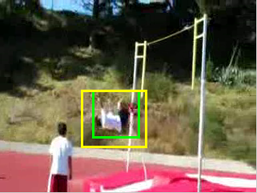}        \smallskip      \\
\includegraphics[height=2.6cm,width=0.98\linewidth]{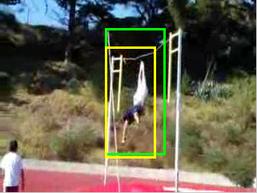}
\centering (g) Pole Vault
\end{minipage}
\hfill
\begin{minipage}{0.24\linewidth}
\begin{center}
\includegraphics[height=2.6cm,width=0.98\linewidth]{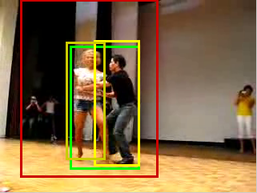}         \smallskip \\
\includegraphics[height=2.6cm,width=0.98\linewidth]{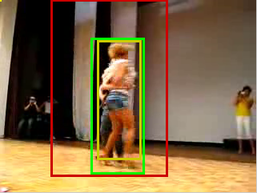}        \smallskip      \\
\includegraphics[height=2.6cm,width=0.98\linewidth]{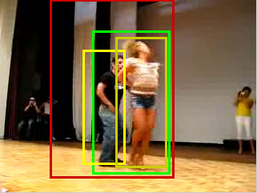}
\centering (h) Salsa Spin
\end{center}
\end{minipage} ~ \smallskip \\
\end{minipage}

\caption{Localization results shown as a sequence of bounding boxes (UCF101): Ground-truth is shown in yellow, correctly localized detections in green and poorly localized ones in red. Caption below each sequence reports the class detected. In case of multiple actors ground-truth boxes are shown darker for the second actor. Poor localization mainly happens in such cases when during the action one of the actors gets occluded (see Salsa Spin) and typically Tubelet often encapsulates both actors together. With 24 classes UCF101 offers larger variety in types of action, which is well captured by our action proposal method.}
\label{tubelet_fig:dets_ucf101}
\end{figure*}

Figure~\ref{tubelet_fig:dets_ucf101} displays some examples of action localizations from UCF101. With 24 classes this dataset offers larger variety in types of actions. Poor localization (shown in red) mainly happens in case of multiple actors, when during the action one of the actors gets occluded (see `Salsa Spin'). Typically, in that case, Tubelets often encapsulates both actors together. However, the varying aspect ratios, diverse locations in the video frames, speed of action and multiple actors are well captured by our action proposal method. 

\section{Conclusions} \label{conclusion}
We presented an unsupervised approach to generate proposals from super-voxels for action localization in videos. This is done by iterative grouping of super-voxels driven by both static features and motion features, motion being the key ingredient. 
We introduced independent motion evidence to characterize how the action related motion deviates from the background. The generated \imemotion maps provide a more efficient alternative for segmentation. Moreover, \imemotion-based features allow for effective and efficient grouping of super-voxels. Our action proposals, Tubelets, are action class independent and implicitly cover variable aspect ratios and temporal lengths. We showed, for the first time, the effectiveness of Tubelets for action localization in \cite{Jain:tubelets}. In this paper, \imemotion maps are presented with further insights and the segmenting \imemotion maps is shown complementary to segmenting input video frames. Additionally, we introduced spatiotemporal refinement and pruning of Tubelets. Spatiotemporal refinement overcomes the tendency of super-voxels to sometimes follow the actor even after the action is completed. This led to improved MABO and Recall scores, especially on the untrimmed videos of MSR-II (Table~\ref{tubelet_table:prune_refine_msr}), while pruning kept the number of Tubelets limited. The impact of these and the other components of Tubelet generation are extensively evaluated in our experiments.

We evaluate our method for both action proposal quality and action localization. For action proposal quality, Tubelets beat all other existing approaches on the three datasets with much fewer number of proposals (Table~\ref{tab:methodsComp}). For action localization, our method leads to the best performance on UCF101 and second best on UCF Sports and MSR-II. The method of \cite{chen2015action} gets best mAP for MSR-II but its performance on UCF Sports is rather modest. Similarly \cite{Weinzaepfel_iccv15} does well on UCF Sports and UCF101 but being supervised in generating proposals is not easy to apply on MSR-II. Ours is the only method that delivers excellent performance on both the trimmed videos of UCF Sports and UCF101 as well as the untrimmed videos of MSR-II.

\bibliographystyle{plainnat}
{\footnotesize
\bibliography{egbib,intro}}

\begin{thebibliography}{72}
\providecommand{\natexlab}[1]{#1}
\providecommand{\url}[1]{\texttt{#1}}
\expandafter\ifx\csname urlstyle\endcsname\relax
  \providecommand{\doi}[1]{doi: #1}\else
  \providecommand{\doi}{doi: \begingroup \urlstyle{rm}\Url}\fi

\bibitem[Brox and Malik(2010)]{BroxJMeccv10}
Thomas Brox and Jitendra Malik.
\newblock Object segmentation by long term analysis of point trajectories.
\newblock In \emph{Proceedings of the European Conference on Computer Vision},
  Sep. 2010.

\bibitem[Cao et~al.(2010)Cao, Liu, and Huang]{Cao:cvpr10}
Liangliang Cao, Zicheng Liu, and Thomas~S. Huang.
\newblock Cross-dataset action detection.
\newblock In \emph{Proceedings of the IEEE Conference on Computer Vision and
  Pattern Recognition}, Jun. 2010.

\bibitem[Chen and Corso(2015)]{chen2015action}
Wei Chen and Jason~J. Corso.
\newblock Action detection by implicit intentional motion clustering.
\newblock In \emph{Proceedings of the IEEE International Conference on Computer
  Vision}, 2015.

\bibitem[Chen et~al.(2014)Chen, Xiong, Xu, and Corso]{chenCVPR14actionness}
Wei Chen, Caiming Xiong, Ran Xu, and Jason Corso.
\newblock Actionness ranking with lattice conditional ordinal random fields.
\newblock In \emph{Proceedings of the IEEE Conference on Computer Vision and
  Pattern Recognition}, pages 748--755, 2014.

\bibitem[Cheron et~al.(2015)Cheron, Laptev, and
  Schmid]{Cheron_2015_ICCVposeCNN}
Guilhem Cheron, Ivan Laptev, and Cordelia Schmid.
\newblock P-cnn: Pose-based cnn features for action recognition.
\newblock In \emph{Proceedings of the IEEE International Conference on Computer
  Vision}, December 2015.

\bibitem[Dalal and Triggs(2005)]{dalal_hog}
Navneet Dalal and Bill Triggs.
\newblock Histograms of oriented gradients for human detection.
\newblock In \emph{Proceedings of the IEEE Conference on Computer Vision and
  Pattern Recognition}, Jun. 2005.

\bibitem[Delaitre et~al.(2010)Delaitre, Laptev, and
  Sivic]{delaitreBMVC2010bowAndParts}
Vincent Delaitre, Ivan Laptev, and Josef Sivic.
\newblock Recognizing human actions in still images: a study of bag-of-features
  and part-based representations.
\newblock In \emph{BMVC}, 2010.

\bibitem[Derpanis et~al.(2013)Derpanis, Sizintsev, Cannons, and
  Wildes]{derpanisPAMI2013actionSpotting}
Konstantinos~G Derpanis, Mikhail Sizintsev, Kevin~J Cannons, and Richard~P
  Wildes.
\newblock Action spotting and recognition based on a spatiotemporal orientation
  analysis.
\newblock \emph{Pattern Analysis and Machine Intelligence, IEEE Transactions
  on}, 35\penalty0 (3):\penalty0 527--540, 2013.

\bibitem[Dollar et~al.(2005)Dollar, Rabaud, Cottrell, and
  Belongie]{dollar_vs-pets}
Piotr Dollar, Vincent Rabaud, Garrison Cottrell, and Serge Belongie.
\newblock Behavior recognition via sparse spatio-temporal features.
\newblock In \emph{Visual Surveillance and Performance Evaluation of Tracking
  and Surveillance}, Oct. 2005.

\bibitem[Everts et~al.(2014)Everts, van Gemert, and
  Gevers]{evertsTIP14evaluation}
Ivo Everts, Jan~C. van Gemert, and Theo Gevers.
\newblock Evaluation of color spatio-temporal interest points for human action
  recognition.
\newblock \emph{{\sc IEEE} Transactions on Image Processing}, 23\penalty0
  (4):\penalty0 1569--1580, 2014.

\bibitem[Felzenszwalb and Huttenlocher(2004)]{Felzenszwalb2004}
Pedro~F. Felzenszwalb and Daniel~P. Huttenlocher.
\newblock Efficient graph-based image segmentation.
\newblock \emph{International Journal of Computer Vision}, 59\penalty0
  (2):\penalty0 167--181, 2004.

\bibitem[Felzenszwalb et~al.(2010)Felzenszwalb, Girshick, McAllester, and
  Ramanan]{Felzenszwalb:DPM}
Pedro~F. Felzenszwalb, Ross~B. Girshick, David~A. McAllester, and Deva Ramanan.
\newblock Object detection with discriminatively trained part-based models.
\newblock \emph{{\sc IEEE} Transactions on Pattern Analysis and Machine
  Intelligence}, 32\penalty0 (9):\penalty0 1627--1645, 2010.

\bibitem[Girshick et~al.(2016)Girshick, Donahue, Darrell, and
  Malik]{Girshick2016}
Ross~B. Girshick, Jeff Donahue, Trevor Darrell, and Jitendra Malik.
\newblock Region-based convolutional networks for accurate object detection and
  semantic segmentation.
\newblock \emph{IEEE Transactions on Pattern Analysis and Machine
  Intelligence}, 38\penalty0 (1):\penalty0 142--158, 2016.

\bibitem[Gkioxari and Malik(2015)]{gkioxariCVPR15actionTubes}
Georgi Gkioxari and Jitendra Malik.
\newblock Finding action tubes.
\newblock In \emph{Proceedings of the IEEE Conference on Computer Vision and
  Pattern Recognition}, 2015.

\bibitem[Hosang et~al.(2016)Hosang, Benenson, Doll{\'a}r, and
  Schiele]{Hosang2016}
Jan Hosang, Rodrigo Benenson, Piotr Doll{\'a}r, and Bernt Schiele.
\newblock What makes for effective detection proposals?
\newblock \emph{IEEE Transactions on Pattern Analysis and Machine
  Intelligence}, 38\penalty0 (4):\penalty0 814--830, 2016.

\bibitem[Huber(1981)]{Huber81}
Peter~J. Huber.
\newblock \emph{{Robust statistics}}.
\newblock Wiley, New York, 1981.

\bibitem[Jain et~al.(2013)Jain, J{\'e}gou, and Bouthemy]{Jain:wflow}
Mihir Jain, Herv{\'e} J{\'e}gou, and Patrick Bouthemy.
\newblock Better exploiting motion for better action recognition.
\newblock In \emph{Proceedings of the IEEE Conference on Computer Vision and
  Pattern Recognition}, Jun. 2013.

\bibitem[Jain et~al.(2014)Jain, van Gemert, J{\'e}gou, Bouthemy, and
  Snoek]{Jain:tubelets}
Mihir Jain, Jan~C. van Gemert, Herv{\'e} J{\'e}gou, Patrick Bouthemy, and Cees
  G.~M. Snoek.
\newblock Action localization by tubelets from motion.
\newblock In \emph{Proceedings of the IEEE Conference on Computer Vision and
  Pattern Recognition}, Jun. 2014.

\bibitem[Jain et~al.(2015{\natexlab{a}})Jain, van Gemert, Mensink, and
  Snoek]{jain2015objects2action}
Mihir Jain, Jan~C van Gemert, Thomas Mensink, and Cees~GM Snoek.
\newblock Objects2action: Classifying and localizing actions without any video
  example.
\newblock In \emph{Proceedings of the IEEE International Conference on Computer
  Vision}, pages 4588--4596, 2015{\natexlab{a}}.

\bibitem[Jain et~al.(2015{\natexlab{b}})Jain, van Gemert, and
  Snoek]{Jain:15kobjects}
Mihir Jain, Jan~C. van Gemert, and Cees G.~M. Snoek.
\newblock What do 15,000 object categories tell us about classifying and
  localizing actions?
\newblock In \emph{Proceedings of the IEEE Conference on Computer Vision and
  Pattern Recognition}, Jun. 2015{\natexlab{b}}.

\bibitem[Jain et~al.(2016)Jain, J{\'e}gou, and Bouthemy]{jain2016improved}
Mihir Jain, Herv{\'e} J{\'e}gou, and Patrick Bouthemy.
\newblock Improved motion description for action classification.
\newblock \emph{Frontiers in ICT}, 2:\penalty0 28, 2016.

\bibitem[J\'egou et~al.(2012)J\'egou, Perronnin, Douze, S\'anchez, P\'erez, and
  Schmid]{JPDSPS12}
Herv\'e J\'egou, Florent Perronnin, Matthijs Douze, Jorge S\'anchez, Patrick
  P\'erez, and Cordelia Schmid.
\newblock Aggregating local descriptors into compact codes.
\newblock \emph{{\sc IEEE} Transactions on Pattern Analysis and Machine
  Intelligence}, 34\penalty0 (9):\penalty0 1704--1716, 2012.

\bibitem[Jhuang et~al.(2013)Jhuang, Gall, Zuffi, Schmid, and
  Black]{Jhuang_2013_ICCVtowardsUnderstandAR}
Hueihan Jhuang, Juergen Gall, Silvia Zuffi, Cordelia Schmid, and Michael~J.
  Black.
\newblock Towards understanding action recognition.
\newblock In \emph{Proceedings of the IEEE International Conference on Computer
  Vision}, December 2013.

\bibitem[Ke et~al.(2005)Ke, Sukthankar, and Hebert]{yanke_iccv05}
Yan Ke, Rahul Sukthankar, and Martial Hebert.
\newblock Efficient visual event detection using volumetric features.
\newblock In \emph{Proceedings of the IEEE International Conference on Computer
  Vision}, Oct. 2005.

\bibitem[Kim and Pavlovic(2010)]{kimECCV2010ordinalRegression}
Minyoung Kim and Vladimir Pavlovic.
\newblock Structured output ordinal regression for dynamic facial emotion
  intensity prediction.
\newblock In \emph{European Conference on Computer Vision}, pages 649--662.
  Springer, 2010.

\bibitem[Kl{\"a}ser et~al.(2008)Kl{\"a}ser, Marszalek, and Schmid]{hog3D}
Alexander Kl{\"a}ser, Marcin Marszalek, and Cordelia Schmid.
\newblock A spatio-temporal descriptor based on 3d-gradients.
\newblock In \emph{Proceedings of the British Machine Vision Conference}, Sep.
  2008.

\bibitem[Kl{\"a}ser et~al.(2012)Kl{\"a}ser, Marsza{\l}ek, Schmid, and
  Zisserman]{klaser2012human}
Alexander Kl{\"a}ser, Marcin Marsza{\l}ek, Cordelia Schmid, and Andrew
  Zisserman.
\newblock Human focused action localization in video.
\newblock In \emph{Trends and Topics in Computer Vision}, pages 219--233, 2012.

\bibitem[Krizhevsky et~al.(2012)Krizhevsky, Sutskever, and
  Hinton]{Krizhevsky_imagenetclassification}
Alex Krizhevsky, Ilya Sutskever, and Geoffrey~E. Hinton.
\newblock Imagenet classification with deep convolutional neural networks.
\newblock In \emph{Advances in Neural Information Processing Systems}, 2012.

\bibitem[Lampert et~al.(2008)Lampert, Blaschko, and Hofmann]{Lampert:ESS}
Christoph~H. Lampert, Matthew~B. Blaschko, and Thomas Hofmann.
\newblock Beyond sliding windows: Object localization by efficient subwindow
  search.
\newblock In \emph{Proceedings of the IEEE Conference on Computer Vision and
  Pattern Recognition}, Jun. 2008.

\bibitem[Lan et~al.(2011)Lan, Wang, and Mori]{tian_iccv11}
Tian Lan, Yang Wang, and Greg Mori.
\newblock Discriminative figure-centric models for joint action localization
  and recognition.
\newblock In \emph{Proceedings of the IEEE International Conference on Computer
  Vision}, Nov. 2011.

\bibitem[Laptev(2005)]{Laptev2005}
Ivan Laptev.
\newblock On space-time interest points.
\newblock \emph{International Journal of Computer Vision}, 64\penalty0
  (2):\penalty0 107--123, 2005.

\bibitem[Ma et~al.(2013)Ma, Zhang, Ikizler-Cinbis, and
  Sclaroff]{ma2013ICCVspacetimesegments}
Shugao Ma, Jianming Zhang, Nazli Ikizler-Cinbis, and Stan Sclaroff.
\newblock Action recognition and localization by hierarchical space-time
  segments.
\newblock In \emph{Proceedings of the IEEE International Conference on Computer
  Vision}, pages 2744--2751, 2013.

\bibitem[Maji et~al.(2011)Maji, Bourdev, and Malik]{majiCVPR2011actionPose}
Subhransu Maji, Lubomir Bourdev, and Jitendra Malik.
\newblock Action recognition from a distributed representation of pose and
  appearance.
\newblock In \emph{Proceedings of the IEEE Conference on Computer Vision and
  Pattern Recognition}, pages 3177--3184, 2011.

\bibitem[Manen et~al.(2013)Manen, Guillaumin, and
  Van~Gool]{manenICCV13objectProposals}
S.~Manen, M.~Guillaumin, and L.~Van~Gool.
\newblock {Prime Object Proposals with Randomized Prim's Algorithm}.
\newblock In \emph{Proceedings of the IEEE International Conference on Computer
  Vision}, 2013.

\bibitem[Mosabbeb et~al.(2014)Mosabbeb, Cabral, De~la Torre, and
  Fathy]{mosabbebACCV2014weakly}
Ehsan~Adeli Mosabbeb, Ricardo Cabral, Fernando De~la Torre, and Mahmood Fathy.
\newblock Multi-label discriminative weakly-supervised human activity
  recognition and localization.
\newblock In \emph{ACCV}, 2014.

\bibitem[Ng et~al.(2015)Ng, Hausknecht, Vijayanarasimhan, Vinyals, Monga, and
  Toderici]{yueCVPR15beyondShortSnippets}
Joe Yue-Hei Ng, Matthew Hausknecht, Sudheendra Vijayanarasimhan, Oriol Vinyals,
  Rajat Monga, and George Toderici.
\newblock Beyond short snippets: Deep networks for video classification.
\newblock In \emph{Proceedings of the IEEE Conference on Computer Vision and
  Pattern Recognition}, pages 4694--4702, 2015.

\bibitem[Odobez and Bouthemy(1995)]{Odobez95robustmultiresolution}
Jean-Marc Odobez and Patrick Bouthemy.
\newblock Robust multiresolution estimation of parametric motion models.
\newblock \emph{Journal of Visual Communication and Image Representation},
  6\penalty0 (4):\penalty0 348--365, Dec. 1995.

\bibitem[Oneata et~al.(2013)Oneata, Verbeek, and Schmid]{Oneata:iccv13}
Dan Oneata, Jakob Verbeek, and Cordelia Schmid.
\newblock {Action and Event Recognition with Fisher Vectors on a Compact
  Feature Set}.
\newblock In \emph{Proceedings of the IEEE International Conference on Computer
  Vision}, Dec. 2013.

\bibitem[Oneata et~al.(2014{\natexlab{a}})Oneata, Revaud, Verbeek, and
  Schmid]{oneataECCV14spatemprops}
Dan Oneata, Jerome Revaud, Jakob Verbeek, and Cordelia Schmid.
\newblock Spatio-temporal object detection proposals.
\newblock In \emph{Proceedings of the European Conference on Computer Vision},
  2014{\natexlab{a}}.

\bibitem[Oneata et~al.(2014{\natexlab{b}})Oneata, Verbeek, and
  Schmid]{oneata:cvpr_14}
Dan Oneata, Jakob Verbeek, and Cordelia Schmid.
\newblock {Efficient Action Localization with Approximately Normalized Fisher
  Vectors}.
\newblock In \emph{Proceedings of the IEEE Conference on Computer Vision and
  Pattern Recognition}, 2014{\natexlab{b}}.

\bibitem[Perronnin and Dance(2007)]{PD07}
Florent Perronnin and Christopher~R. Dance.
\newblock Fisher kernels on visual vocabularies for image categorization.
\newblock In \emph{Proceedings of the IEEE Conference on Computer Vision and
  Pattern Recognition}, 2007.

\bibitem[Perronnin et~al.(2010)Perronnin, S\'anchez, and Mensink]{PSM10}
Florent Perronnin, Jorge S\'anchez, and Thomas Mensink.
\newblock Improving the fisher kernel for large-scale image classification.
\newblock In \emph{Proceedings of the European Conference on Computer Vision},
  Sep. 2010.

\bibitem[Piriou et~al.(2006)Piriou, Bouthemy, and Yao]{Piriou_ieee-tip2006}
Gwenaëlle Piriou, Patrick Bouthemy, and Jian-Feng Yao.
\newblock Recognition of dynamic video contents with global probabilistic
  models of visual motion.
\newblock \emph{{\sc IEEE} Transactions on Image Processing}, 15\penalty0
  (11):\penalty0 3417--3430, 2006.

\bibitem[Puscas et~al.(2015)Puscas, Sangineto, Culibrk, and
  Sebe]{marianICCV2015unsupervisedTube}
Mihai Puscas, Enver Sangineto, Dubravko Culibrk, and Nicu Sebe.
\newblock Unsupervised tube extraction using transductive learning and dense
  trajectories.
\newblock In \emph{Proceedings of the IEEE International Conference on Computer
  Vision}, 2015.

\bibitem[Raptis et~al.(2012)Raptis, Kokkinos, and Soatto]{Raptis:cvpr12}
Michalis Raptis, Iasonas Kokkinos, and Stefano Soatto.
\newblock Discovering discriminative action parts from mid-level video
  representations.
\newblock In \emph{Proceedings of the IEEE Conference on Computer Vision and
  Pattern Recognition}, Jun. 2012.

\bibitem[Rodriguez et~al.(2008)Rodriguez, Ahmed, and Shah]{Rodriguez:cvpr08}
Mikel~D. Rodriguez, Javed Ahmed, and Mubarak Shah.
\newblock Action mach: a spatio-temporal maximum average correlation height
  filter for action recognition.
\newblock In \emph{Proceedings of the IEEE Conference on Computer Vision and
  Pattern Recognition}, Jun. 2008.

\bibitem[S{\'a}nchez et~al.(2013)S{\'a}nchez, Perronnin, Mensink, and
  Verbeek]{Sanchez2013}
Jorge S{\'a}nchez, Florent Perronnin, Thomas Mensink, and Jakob Verbeek.
\newblock Image classification with the fisher vector: Theory and practice.
\newblock \emph{International Journal of Computer Vision}, 105\penalty0
  (3):\penalty0 222--245, 2013.

\bibitem[Sch\"uldt et~al.(2004)Sch\"uldt, Laptev, and Caputo]{kth_icpr04}
Christian Sch\"uldt, Ivan Laptev, and Barbara Caputo.
\newblock Recognizing human actions: A local svm approach.
\newblock In \emph{Proceedings of International Conference of Pattern
  Recognition}, 2004.

\bibitem[Simonyan and Zisserman(2014)]{SimonyanNIPS14}
Karen Simonyan and Andrew Zisserman.
\newblock Two-stream convolutional networks for action recognition in videos.
\newblock In \emph{NIPS}, 2014.

\bibitem[Soomro et~al.(2012)Soomro, Zamir, and Shah]{ucf101}
Khurram Soomro, Amir~Roshan Zamir, and Mubarak Shah.
\newblock {UCF101:} {A} dataset of 101 human actions classes from videos in the
  wild.
\newblock \emph{CoRR}, 2012.
\newblock URL \url{http://arxiv.org/abs/1212.0402}.

\bibitem[Soomro et~al.(2015)Soomro, Idrees, and
  Shah]{soomro2015ICCVcontextWalk}
Khurram Soomro, Haroon Idrees, and Mubarak Shah.
\newblock Action localization in videos through context walk.
\newblock In \emph{Proceedings of the IEEE International Conference on Computer
  Vision}, 2015.

\bibitem[Szegedy et~al.(2015)Szegedy, Liu, Jia, Sermanet, Reed, Anguelov,
  Erhan, Vanhoucke, and Rabinovich]{googlenet}
Christian Szegedy, Wei Liu, Yangqing Jia, Pierre Sermanet, Scott Reed, Dragomir
  Anguelov, Dumitru Erhan, Vincent Vanhoucke, and Andrew Rabinovich.
\newblock Going deeper with convolutions.
\newblock In \emph{Proceedings of the IEEE Conference on Computer Vision and
  Pattern Recognition}, 2015.

\bibitem[Tian et~al.(2013)Tian, Sukthankar, and Shah]{Yicong:sdpm}
Yicong Tian, Rahul Sukthankar, and Mubarak Shah.
\newblock Spatiotemporal deformable part models for action detection.
\newblock In \emph{Proceedings of the IEEE Conference on Computer Vision and
  Pattern Recognition}, Jun. 2013.

\bibitem[Tran and Yuan(2011)]{Tran:cvpr11}
Du~Tran and Junsong Yuan.
\newblock Optimal spatio-temporal path discovery for video event detection.
\newblock In \emph{Proceedings of the IEEE Conference on Computer Vision and
  Pattern Recognition}, Jun. 2011.

\bibitem[Tran and Yuan(2012)]{Tran:nips12}
Du~Tran and Junsong Yuan.
\newblock Max-margin structured output regression for spatio-temporal action
  localization.
\newblock In \emph{Advances in Neural Information Processing Systems}, Dec.
  2012.

\bibitem[Tran et~al.(2015)Tran, Bourdev, Fergus, Torresani, and
  Paluri]{tranICCV2015c3d}
Du~Tran, Lubomir Bourdev, Rob Fergus, Lorenzo Torresani, and Manohar Paluri.
\newblock Learning spatiotemporal features with 3d convolutional networks.
\newblock In \emph{Proceedings of the IEEE International Conference on Computer
  Vision}, pages 4489--4497, 2015.

\bibitem[Uijlings et~al.(2013)Uijlings, van~de Sande, Gevers, and
  Smeulders]{Jasper:selective}
Jasper R.~R. Uijlings, Koen. E.~A. van~de Sande, Theo Gevers, and Arnold W.~M.
  Smeulders.
\newblock Selective search for object recognition.
\newblock \emph{International Journal of Computer Vision}, 104\penalty0
  (2):\penalty0 154--171, 2013.

\bibitem[van~de Sande et~al.(2014)van~de Sande, Snoek, and
  Smeulders]{SandeCVPR14}
Koen E.~A. van~de Sande, Cees G.~M. Snoek, and Arnold W.~M. Smeulders.
\newblock Fisher and vlad with flair.
\newblock In \emph{Proceedings of the IEEE Conference on Computer Vision and
  Pattern Recognition}, 2014.

\bibitem[van Gemert et~al.(2015)van Gemert, Jain, Gati, and
  Snoek]{Gemert_BMVC15}
Jan~C. van Gemert, Mihir Jain, Ella Gati, and Cees G.~M. Snoek.
\newblock {APT}: Action localization proposals from dense trajectories.
\newblock In \emph{Proceedings of the British Machine Vision Conference}, 2015.

\bibitem[Viola and Jones(2004)]{violajones_ijcv}
Paul~A. Viola and Michael~J. Jones.
\newblock Robust real-time face detection.
\newblock \emph{International Journal of Computer Vision}, 57\penalty0
  (2):\penalty0 137--154, 2004.

\bibitem[Wang and Schmid(2013)]{wang:imptraj13}
Heng Wang and Cordelia Schmid.
\newblock {Action Recognition with Improved Trajectories}.
\newblock In \emph{Proceedings of the IEEE International Conference on Computer
  Vision}, Dec. 2013.

\bibitem[Wang et~al.(2011)Wang, Kl{\"a}ser, Schmid, and Liu]{densetrack11}
Heng Wang, Alexander Kl{\"a}ser, Cordelia Schmid, and Cheng-Lin Liu.
\newblock Action recognition by dense trajectories.
\newblock In \emph{Proceedings of the IEEE Conference on Computer Vision and
  Pattern Recognition}, Jun. 2011.

\bibitem[Wang et~al.(2013)Wang, Kl{\"a}ser, Schmid, and Liu]{wang:ijcv13}
Heng Wang, Alexander Kl{\"a}ser, Cordelia Schmid, and Cheng-Lin Liu.
\newblock {Dense trajectories and motion boundary descriptors for action
  recognition}.
\newblock \emph{International Journal of Computer Vision}, 103\penalty0
  (1):\penalty0 60--79, 2013.

\bibitem[Wang et~al.(2015{\natexlab{a}})Wang, Oneata, Verbeek, and
  Schmid]{Wang2015}
Heng Wang, Dan Oneata, Jakob Verbeek, and Cordelia Schmid.
\newblock A robust and efficient video representation for action recognition.
\newblock \emph{International Journal of Computer Vision}, pages 1--20,
  2015{\natexlab{a}}.

\bibitem[Wang et~al.(2014)Wang, Qiao, and Tang]{Wang_dynamicPoslets}
Limin Wang, Yu~Qiao, and Xiaoou Tang.
\newblock Video action detection with relational dynamic-poselets.
\newblock In \emph{Proceedings of the European Conference on Computer Vision},
  2014.

\bibitem[Wang et~al.(2015{\natexlab{b}})Wang, Qiao, and Tang]{wangCVPR15tdd}
Limin Wang, Yu~Qiao, and Xiaoou Tang.
\newblock Action recognition with trajectory-pooled deep-convolutional
  descriptors.
\newblock In \emph{Proceedings of the IEEE Conference on Computer Vision and
  Pattern Recognition}, pages 4305--4314, 2015{\natexlab{b}}.

\bibitem[Wang and Mori(2011)]{wangPAMI2011hiddenPart}
Yang Wang and Greg Mori.
\newblock Hidden part models for human action recognition: Probabilistic versus
  max margin.
\newblock \emph{IEEE Transactions on Pattern Analysis and Machine
  Intelligence}, 33\penalty0 (7):\penalty0 1310--1323, 2011.

\bibitem[Weinzaepfel et~al.(2015)Weinzaepfel, Harchaoui, and
  Schmid]{Weinzaepfel_iccv15}
Philippe Weinzaepfel, Zaid Harchaoui, and Cordelia Schmid.
\newblock Learning to track for spatio-temporal action localization.
\newblock In \emph{Proceedings of the IEEE International Conference on Computer
  Vision}, 2015.

\bibitem[Xu and Corso(2012)]{xuCVPR12SuperVoxelEvaluation}
Chenliang Xu and Jason~J. Corso.
\newblock Evaluation of super-voxel methods for early video processing.
\newblock In \emph{Proceedings of the IEEE Conference on Computer Vision and
  Pattern Recognition}, 2012.

\bibitem[Yu and Yuan(2015)]{yuCVPR15fasttubes}
Gang Yu and Junsong Yuan.
\newblock Fast action proposals for human action detection and search.
\newblock In \emph{Proceedings of the IEEE Conference on Computer Vision and
  Pattern Recognition}, 2015.

\bibitem[Yuan et~al.(2009)Yuan, Liu, and Wu]{yuan_cvpr09}
Junsong Yuan, Zicheng Liu, and Ying Wu.
\newblock Discriminative subvolume search for efficient action detection.
\newblock In \emph{Proceedings of the IEEE Conference on Computer Vision and
  Pattern Recognition}, Jun. 2009.

\bibitem[Yuan et~al.(2011)Yuan, Liu, and Wu]{Yuan:pami11}
Junsong Yuan, Zicheng Liu, and Ying Wu.
\newblock Discriminative video pattern search for efficient action detection.
\newblock \emph{{\sc IEEE} Transactions on Pattern Analysis and Machine
  Intelligence}, 33\penalty0 (9):\penalty0 1728--1743, 2011.

\end{thebibliography}

\end{document}